# Early warning signals for loss of control

Jasper J. van Beers[1]*, Marten Scheffer[2], Prashant Solanki[1], Ingrid A. van de Leemput[2], Egbert H. van Nes[2], and Coen C. de Visser[1]

**Abstract**

Maintaining stability in feedback systems, from aircraft and autonomous robots to biological and physiological systems, relies on monitoring their behavior and continuously adjusting their inputs. Incremental damage can make such control fragile. This tends to go unnoticed until a small perturbation induces instability (i.e. loss of control). Traditional methods in the field of engineering rely on accurate system models to compute a safe set of operating instructions, which become invalid when the, possibly damaged, system diverges from its model. Here we demonstrate that the approach of such a feedback system towards instability can nonetheless be monitored through dynamical indicators of resilience. This holistic system safety monitor does not rely on a system model and is based on the generic phenomenon of critical slowing down, shown to occur in the climate, biology and other complex nonlinear systems approaching criticality. Our findings for engineered devices opens up a wide range of applications involving real-time early warning systems as well as an empirical guidance of resilient system design exploration, or 'tinkering'. While we demonstrate the validity using drones, the generic nature of the underlying principles suggest that these indicators could apply across a wider class of controlled systems including reactors, aircraft, and self-driving cars.

## I. Main

CONTROLLED systems are deeply intertwined with our daily lives, from power generation & distribution[1,2] to autonomous vehicles[3,4,5] but also the human body[6,7]. Across all of these diverse systems, the core objective of the 'controller' is the same. That is, to monitor the system's behavior and guide it to the desired set of states. While this feedback loop typically works well under nominal conditions, any abnormalities or damage incurred by the system can make this feedback relationship fragile[8,9]. In some cases, it leaves the system vulnerable to small perturbations which can induce instability from which it is difficult, or even impossible, for the system to recover in time. We refer to such events as "loss of control".

Consider, for example, the fatal accident of Sriwijaya Air flight SJ-182 which entered an uncontrolled descent shortly after take-off on January 9, 2021[10]. The subsequent accident report[10] concluded that the buildup of faults within the aircraft, coupled with the overcompensation and delayed reaction of the control systems, ultimately resulted in loss of control. Such loss of stability can likewise occur in systems beyond engineering, such as the decline in postural balance as humans age[11] or even the human body as a whole, wherein cascading failure of several degraded subsystems can lead to a catastrophic system-wide collapse[12]. Analogously, no single component caused the fatal accident of Sriwijaya Air SJ-182 alone; instead, it was the interconnected system as a whole that failed.

Avoiding such accidents has long motivated innovation in the field of engineering[13,14]. Recognizing that real-world systems are uncertain, the field of robust control pursues the design of controllers which guarantee safety and performance despite model and controller uncertainties[15,16,17]. Similarly, much work is being done to design specialized fault-tolerant controllers that can maintain safety across an assortment of fault scenarios[9,18,19]. Another popular view on safety is to consider the system's capabilities through reachability analysis[20,21]. The goal is to establish a region of attraction for a safe set of states to avoid collisions[22,23,24] or to remain within a prescribed safe operational envelope[25,26,27]. Advancements in these fields have improved safety across many engineered systems. However, these control theoretic approaches all share a key limitation in that they rely on (a collection of) system models to establish safety. When these models become misaligned with reality, even well-engineered systems can suffer loss of control.

An alternative solution is therefore to find ways of anticipating loss of control that do not rely on a pre-existing system model. Here we develop such a model-free approach, based on the fact that feedback instability can be considered a critical transition. This allows for its prediction via generic indicators of resilience[28,29,30]. Using drones, we show that a controlled system's proximity to loss of control can indeed be monitored holistically, without the need for its system model, through the generic phenomenon of critical slowing down.

---

[1]These authors are with the Faculty of Aerospace Engineering, Delft University of Technology, 2629 HS, Delft, the Netherlands.

[2]These authors are with the Aquatic Ecology and Water Quality Management group, Wageningen University and Research, 6708 PB, Wageningen, the Netherlands.

*Corresponding author

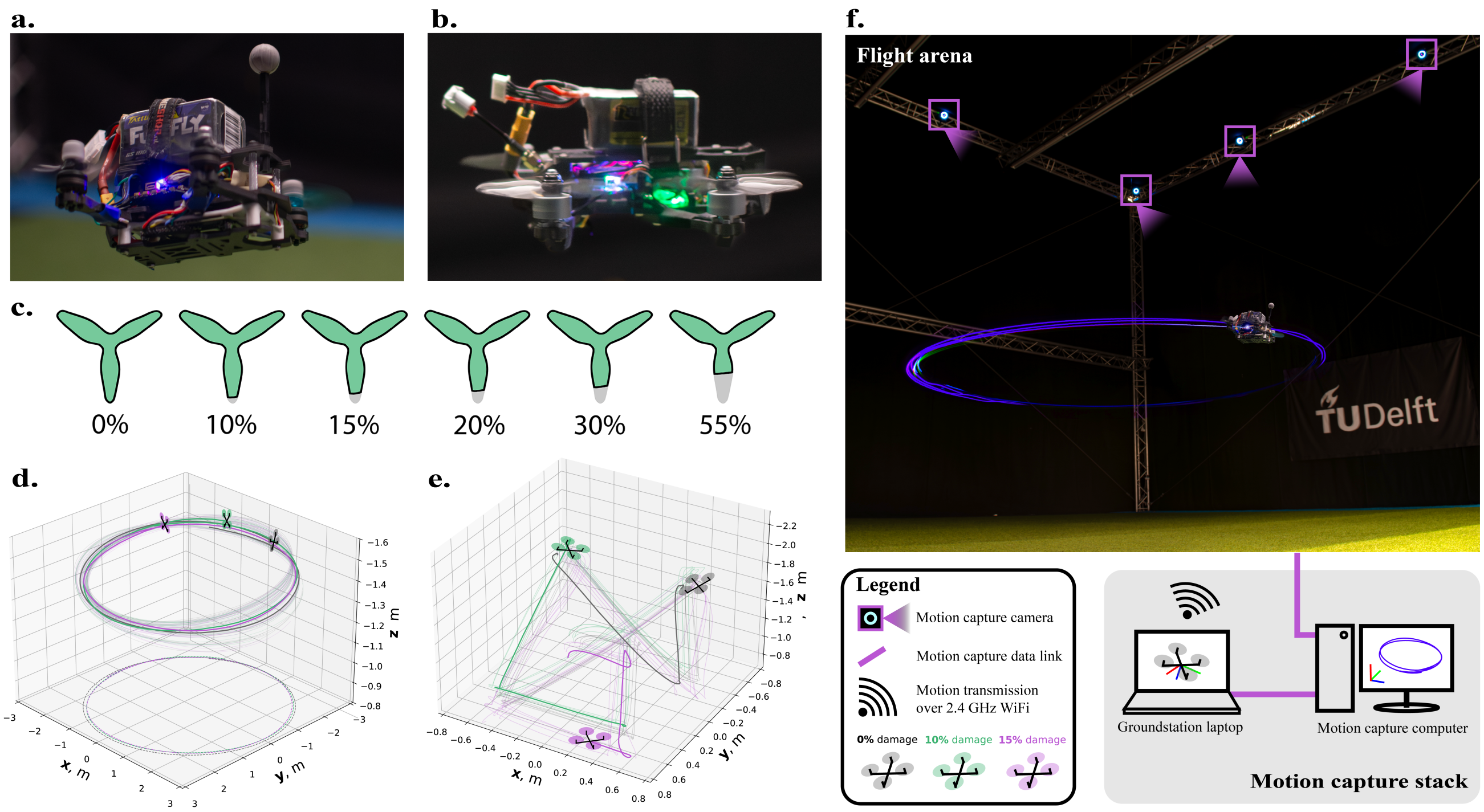

**Figure 1 | Quadrotor flight experiments with incremental blade damage.** We conduct experiments with two quadrotors: the autonomous DragonFly (**a**) controlled using the *INDIFlight* controller and the human-piloted HoverFly (**b**) operating the *Betaflight* controller. Flight experiments are conducted with various degrees of blade damage (**c**) from healthy (0% damage) to 55% blade tip damage. We task the DragonFly with autonomous hovering and trajectory tracking tasks (see **d** for forward circular flight and **e** for tracking the vertices of a 3-dimensional box in position; the opaque lines indicate independent flights while the solid lines depict a portion of the flight trajectory at various discrete damage levels). Conversely, the non-autonomous HoverFly is only flown in hover. The autonomous flights of the DragonFly are facilitated by a motion capture system (OptiTrack) which observes the quadrotor state and relays these measurements via WiFi on a groundstation laptop (**f**).

Critical slowing down has been shown to occur in ecological[31,32,33], biological[34,35,36], and many other complex dynamical systems[37,38,39]. This is most commonly done by determining the lag-1 autocorrelation and variance in a moving window along timeseries of the system states. Moreover, it has recently been shown that critical slowing down can be measured in bursts: brief periods of measurement in intervals preceding a critical transition[40]. Similarly, we monitor the step-wise deterioration in resilience of a quadrotor (a four propeller helicopter) subject to propeller damage increased in increments up to the point of instability. We show that the quadrotor can tolerate blade damage in some operating conditions but loses control completely in others. Therefore, these flight experiments serve as a proxy for situations where gradual damage can be incurred by a controlled system, nudging it closer to instability, even if it remains deceptively operational. This insidious approach towards instability is reflected by the dynamical indicators of resilience.

## II. Generic monitoring of loss of control

Many controllers rely on feedback information to guide a system towards the desired behavior. Such a system is 'closed-loop' in the sense that the controller relies on state information (measurements) to facilitate control. The characteristics of this feedback information thus affect the resilience of the controller. For instance, our ability to maintain postural balance benefits from visual feedback and any impairments (e.g. closed-eyes) inhibit this ability[11]. In engineered systems, the feedback signal is often imperfect, containing sensor dynamics, measurement delays, and inaccuracies (perturbations) such as bias and noise. These imperfections misinform the controller and can thus lead to feedback instability. Box 1 and Box 2 provide illustrative examples of how such instability can arise in controlled systems. In control theory, we refer to this as 'closed-loop instability'; the system in combination with the controller is unstable.

The issue of instability becomes especially problematic for controlled systems which not only suffer from measurement imperfections but also endure gradual changes to the system dynamics. Take, for instance, natural degradation due to wear and tear, or component damage itself, or even (software) system upgrades & improvements. The consequences of these changes on stability can go unnoticed as the controller seeks to maintain the controlled equilibrium until an imminent instability suddenly becomes apparent.

There is growing evidence that, across wildly different systems, such feedback instability can be predicted through critical slowing down (CSD)[29,30,36,41]. However, current demonstrations for engineered systems[29,30] do not consider the influence of a controller that actively drives the system. Due to the persistent feedback influence of the controller, alongside the mechanisms which govern the inherent stability of the system itself, it is often difficult to generically ascertain the exact moment of instability through CSD alone. Thus, rather than predicting this exact moment in the conventional way, we instead apply the generic indicators of resilience to monitor *shifts* in stability for the *current* controlled system as a whole. This nuanced difference affords us insights into a deterioration in controlled system resilience without knowing the exact mechanisms that culminate in this nudge towards instability. In fact, the generic indicators of resilience are consistent with traditional *system model dependent* control theoretic methods for assessing stability, robustness, and resilience (shown in Supplementary Information Section VII). In summary, also in a controlled system we can expect the characteristic symptoms of CSD as it approaches instability.

### *A. Quadrotors as an example*

We apply these dynamical indicators of resilience to monitor shifts in stability of quadrotors (fig. 1 **a** and **b**) subject to incremental propeller damage (fig. 1 **c**) up to the point of instability. The quadrotor can be regarded as a good example of a complex feedback system; its dynamics are highly nonlinear and it is subject to many disturbances that are difficult to predict or measure, such as aerodynamic interactions and component manufacturing inconsistencies. Nevertheless, agile flight is afforded by clever engineering that mitigates many of the challenges associated with controlling the quadrotor. For example, an array of targeted (and often adaptive) feedback filters limit the severity of vibrations and sensor noise generated by the aerodynamics of the propellers and surrounding airflow[42,43]. Such effects would otherwise destabilize the controller[44] through self-reinforcing feedback loops (as described in Box 2). However, the use of these filters comes at the cost of additional feedback delays which can also degrade system stability (see Box 1).

Clearly, such aggressive engineering holds the quadrotor in a delicate dance of stability. Even small changes to the system, such as propeller damage, can induce loss of control. Not only does this damage limit the capability of the affected propeller, it also generates additional (structural) dynamics and asymmetrically intensifies the already problematic vibrations and sensor noise[45,46,47]. These effects need to be compensated for by the remaining (healthy) rotors. In tandem, the targeted filters designed to limit these issues become less effective, allowing (some of) the problematic dynamics to contaminate the feedback signal. While further clever engineering may in principle accommodate such effects[18,48], this requires model knowledge that is lacking in the face of unknown system changes.

### *B. Quadrotor flight experiments*

The adverse effects of propeller damage can result in 'flyaway' loss of control events (i.e. uncontrolled ascending flight provoked by rapid and unstable oscillations in rotor speeds, see fig. 2 **a** and Supplementary Video 2). These flyaways occur consistently near a critical damage level of 30% (see Supplementary Video 1) for our autonomous *INDIFlight* controlled quadrotor, named 'DragonFly' (fig. 1 **a**). Though the *INDIFlight* controller can maintain stability below the critical 30% damage threshold, loss of control occasionally occurs already at 15% damage in demanding operating conditions (such as fig. 2 **a**).

Therefore, to monitor this approach towards instability, we conduct flight experiments with the DragonFly subject to 0%, 10% and 15% blade damage levels (see fig. 1 **c**). These flight experiments involve both hovering and trajectory tracking tasks (fig. 3) in either windless or windy conditions to reflect real-world operations (more details in section IV-A). Furthermore, to limit coincidence, we conducted multiple flights at each damage level and damage location combination. All these flights (n = 247) are summarized in table II.

For each flight, we quantify CSD through the lag-1 autocorrelation metric (AC1)[28,49] along a sliding window applied to the rotor speed measurement timeseries. The AC1 are calculated over a sliding window to form a persistent monitor of instability suitable for real-world operations. Though we explicitly damage the quadrotor here, in reality, damage may accumulate during operation or between missions (e.g. during transportation). Here, the AC1 timeseries are calculated for each of the (detrended) rotor speed measurements individually in order to observe the holistic (i.e. net) effect of the damage (more information on the generic indicator construction process can be found in section IV-B and fig. 4).

These early warning indicators reveal how the quadrotor's loss of stability as the blade damage grows is reflected by subtle changes in dynamics consistent with CSD. Aggregating the early warning indicators (i.e. the AC1 values) by damage level across flight tasks shows a clear increase in AC1 alongside the degree of blade damage from (nominal) 0% damage to 15% damage (fig. 2 **b**). This trend is consistent across all damage locations. These observations are supported through statistical analyses (detailed in section IV-C), which remain significant despite variations in the generic indicator construction parameters (see fig. 6). Here, we report a few notable results.

A one-sided Brunner Munzel test shows that the increase in median AC1 due to damage, no matter the damage location, is significant at both the 10% ($t = 3.70, p < 0.001$) and 15% ($t = 12.45, p < 0.001$) damage levels (see table III). The estimated probability of superiority (i.e. Vargha Delaney A) is $PS = 0.59$ (*95% CI: [0.54, 0.63]*) at the 10% damage level and $PS = 0.76$ (*95% CI: [0.72, 0.80]*) with 15% damage. This indicates that blade damage consistently and strongly increases the underlying AC1 distribution, which is consistent with the qualitative increase in loss of control events observed at the 15% damage level (e.g. fig. 2 **a** and Supplementary Video 3).

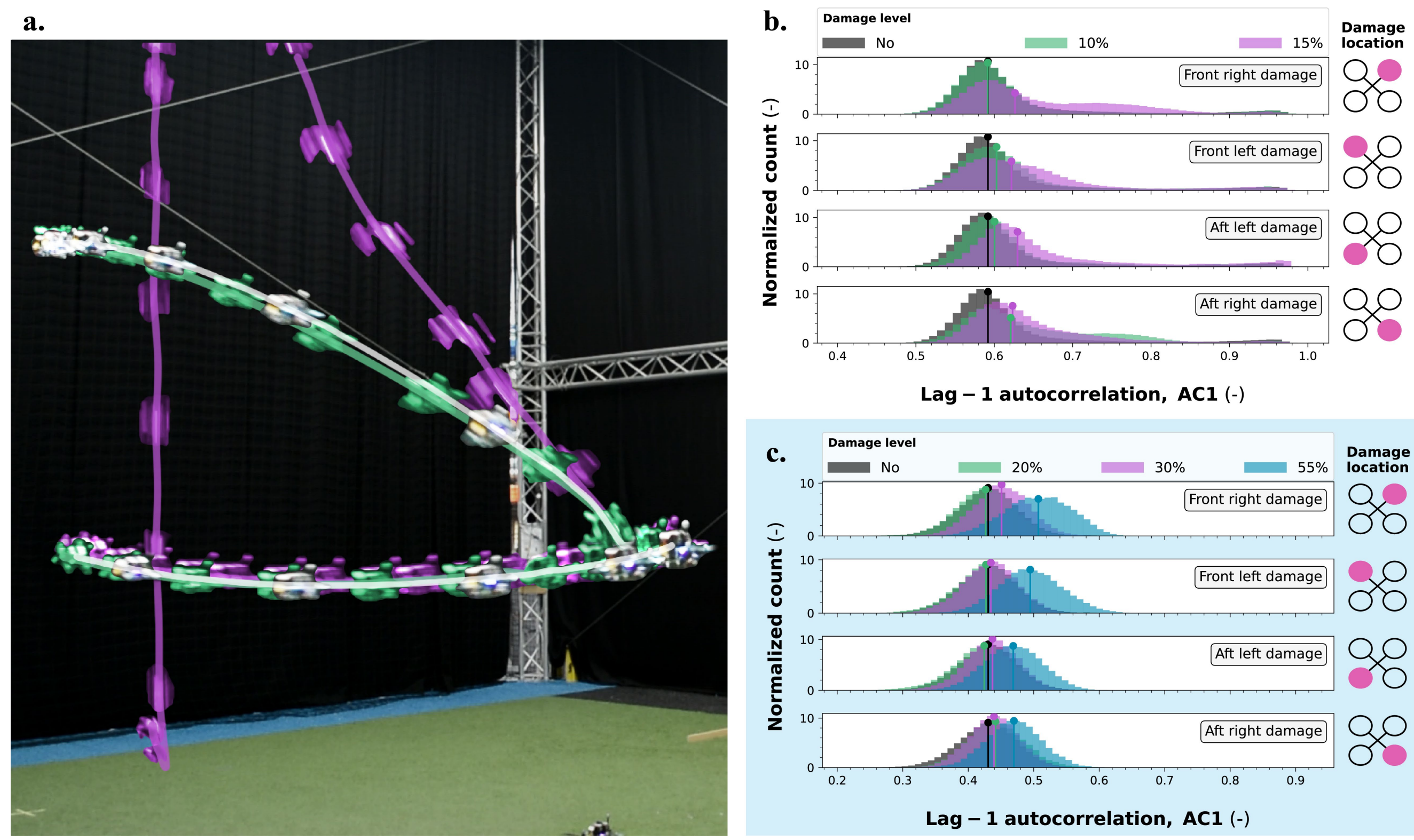

**Figure 2 | Quadrotor instability due to incremental blade damage.** The quadrotor system shifts towards instability (loss of control) with increasing propeller damage. The composite image (**a**) depicts the occurrence of such a loss of control event at the 15% damage level (purple silhouette) during flights of the *INDIFlight* controlled 'DragonFly' quadrotor. The presence of this vulnerability remains hidden in the flight performance at the 10% damage level (green silhouette), which closely follows the nominal 0% damage (white silhouette) flight. Nonetheless, this approach to instability is reflected by generic indicators of critical slowing down, as shown by corresponding increases in lag-1 autocorrelation (AC1) metric alongside the damage severity for the *INDIFlight* controlled DragonFly (**b**). The vertical axis ('Normalized count') denotes the number of AC1 values in each histogram bin, normalized such that the total area under the histogram equals one. For each damage location, the depicted AC1 distributions are composed of the AC1 values combined across all four rotor measurements (i.e. including the undamaged rotors) and flight conditions. Moreover, the characteristic increase in AC1 alongside damage severity is also observed for a similar quadrotor that runs a more (blade) damage resilient controller (*Betaflight 4.3.2*), albeit tolerant of a higher damage degree (**c**). Though the extent of the AC1 increase depends on the location of the damage and controller type, both **b** and **c** show a consistent increase in AC1 alongside the damage level across all damage locations.

Furthermore, though the quadrotor is often considered symmetric, our AC1 statistical analyses reveal hidden asymmetries in its stability. For example, the increase in AC1 due to aft rotor damage is significantly higher than front rotor damage at both the 10% ($t = 3.56, p < 0.001$) and 15% ($t = 2.35, p = 0.001$) damage levels (see table III). Similarly, 15% damage to the left rotors tends to increase the AC1 significantly more than damage to the right rotors: $t = 1.71, p = 0.044$ (see table III).

These curious asymmetries emerge due to system specific nuances, despite a seemingly consistent design. For instance, the front-aft asymmetry is most likely due to the rear placement of the battery on the quadrotor (see fig. 1 **a**), which pulls the center of gravity backward. This demands more effort from the two aft rotors. Likewise, the left-right asymmetry is a function of the varying construction quality of the motors and propellers[46,47]. On the DragonFly, the aft left motor runs 5°C warmer than the other motors, indicating a fault or damage (see Supplementary Video 3). Such characteristic system traits have a clear effect on system stability but are challenging, perhaps even impossible, to model effectively. Nonetheless, their net effect on stability can effectively be monitored through our dynamical indicators of resilience.

To further demonstrate this holistic monitoring ability, we conduct additional incremental damage flight experiments with a similar quadrotor (fig. 1 **b**) that runs a more damage tolerant flight control software: *Betaflight* (more details in section IV-A). By forfeiting autonomous capabilities, this controller relies less on the sensors (specifically, the accelerometers) which are most affected by the propeller damage induced oscillations. Consequently, the *Betaflight* controlled quadrotor can hover at damage levels of 55%, unlike the 30% limit of its *INDIFlight* counterpart. This improved robustness is reflected in the associated AC1 distributions (fig. 2 **c**). Although the *INDIFlight* controller may be modified to also improve its resilience to blade damage, doing so simply shifts the problem of instability to a higher damage level. With or without these improvements, the generic indicators discussed here remain sensitive to shifts in stability.

## III. OUTLOOK

Our quadrotor flight experiments demonstrate that critical slowing down (CSD) can be used to detect loss of stability in controlled systems. This opens up the entirely novel possibility to monitor resilience of dynamic systems on the fly, rather than compute it upfront from models that can become inaccurate in the face of incremental damage or other problems. For example, many engineered systems suffer component degradation over their life cycle, may harbor hidden inconsistencies in manufacturing quality, and can behave differently depending on environmental conditions (e.g. dry versus wet road surfaces). A novel generation of controllers based on our findings may be designed to monitor for hidden changes in stability through CSD and leverage this information to warn in much the same way that we perceive pain: if our ankle hurts, we may not know *why* it hurts, but we are aware that we should be careful as something is *different* from normal. Depending on the severity of the shift in stability, the new class of controllers may then advise an intervention to maintain safety. Thus, CSD equips the controller of a feedback system with awareness of changes to its safe operating space, bridging the gap between internal models and reality.

While our approach offers (early) warning of a nearing instability, it does not directly give an indication of how ‘far’ away the critical threshold is. Likewise, the source(s) of the instability remains unknown. Therefore, it is challenging to determine from the indicator alone how to prevent the loss of stability. Moreover, our dynamic indicator is a relative metric in the sense that we require a reference for nominal system behavior. Therefore, if the nominal system itself is already dangerously close to instability, then the system may reach its critical threshold before any meaningful shifts in stability can be observed.

Furthermore, while our approach is generic, it relies on measuring informative variables to observe meaningful changes in stability. Identifying such variables remains an open challenge for practical use in many real systems. In general, we view variables which relate to the output of the controller and input to the ‘uncontrolled’ system as prime candidates (take, for example, the rotor speed measurements for the quadrotor). Even so, some systems may be limited in their observability of these variables, especially at the necessary fidelity (i.e. accuracy and sampling rate).

Despite such limitations, the feasibility of real-time monitoring of the inherent stability of engineered systems demonstrated here has profound practical implications. Those include the possibility to warn for dangerous instability, prompting operators to halt use of the system for inspection and maintenance. Likewise, these indicators may be used in a quality control setting during production and manufacturing to identify anomalies. An entirely different application of our approach is to use it as a design tool allowing one to ‘tinker’ with engineered systems and measure immediately whether small adjustments are an improvement when it comes to stability. More speculative, automated resilience monitoring may allow systems to diagnose issues during operation and guide real-time reconfiguration in a way that improves resilience thus adapting to changing environmental conditions or system damage. To this end, we may draw inspiration from the natural world to explore and guide such real-time adaptive behavior in engineered systems.

## MAIN REFERENCES

## IV. Methods

### A. Quadrotor flight data collection

Flight experiments are conducted with two quadrotors built using similar off-the-shelf commercial components (see table I), but operated using two different flight controllers (i.e. software). The main results of incremental blade damage are obtained through the autonomous drone named 'DragonFly' (fig. 1 **a**), which runs the *INDIFlight*[i] controller. Conversely, the 'HoverFly' quadrotor (fig. 1 **b**) operates the non-autonomous *Betaflight 4.3.2*[ii] controller and is used to demonstrate how closed-loop instability is a function of the controller, a trait that is reflected through the generic indicators of instability.

*1) Flight control software:* A near 'open-loop' control policy has been adopted by *Betaflight*, affording flexible flight to hobbyist human pilots. While desirable for drone racing and freestyle flight, the low-level control managed by *Betaflight* leaves the human pilot largely responsible for maintaining stability. Consequently, *Betaflight* does not natively support autonomous flight. In contrast, *INDIFlight* extends the base *Betaflight* infrastructure to enable autonomous flight by relying on additional sensors. Namely, the onboard accelerometer measurements enable thrust control while an external motion capture system (OptiTrack) is used to measure the quadrotor's position, velocity, and orientation (see fig. 1 **f**).

However, it is these additional feedback measurement dependencies that provoke instability within *INDIFlight* when operating with propeller damage. The accelerometer is sensitive to excessive vibrations - which intensify with growing damage - that cause aggressive over-corrections from the controller, leading to instabilities (following the mechanisms described in box 2). Note that although one may design targeted filters to mitigate this sensitivity, doing so only postpones the issue of instability; at a certain damage level, the quadrotor is unable to fly irrespective of the filtering applied.

Moreover, the transmission of the external motion capture data also affects closed-loop stability due to interference and congestion in the 2.4 GHz WiFi communication link between the ground station and quadrotor (see fig. 1 **f**). The result is an inconsistent transmission rate of 100-200 Hz with occasional dropouts. In contrast, the control loop of *INDIFlight* runs onboard at 1000 Hz and therefore holds the position and attitude measurements until a new measurement update is available. This can result in mismatches between the controller's belief of the current state and the true state of the quadrotor.

*2) Autonomous flight experiments:* Autonomous flights of the *INDIFlight* controlled DragonFly consist of both hovering and trajectory tracking tasks with and without propeller damage (see fig. 3). The trajectory tracking tasks involve following a circle of 3 *m* radius at airspeeds of 1.5 m/s & 3.0 m/s and maneuvers between the vertices of a 3-dimensional rectangle in space, designed to excite each of the four motors individually at least once during the flight. Furthermore, some trajectories are flown in both windless and windy conditions to challenge the controller with further perturbations. Here, wind is generated by a large fan (see fig. 5) which disturbs the quadrotor along its flight path. Adequately rejecting this disturbance can become demanding under propeller damage.

The DragonFly is unable to maintain stable flight at a damage level of 30% (Supplementary Video 1). Thus, on approach to this threshold, flight data is collected at damage levels of 0%, 10%, and 15% at each rotor location separately. We do not collect data on simultaneous rotor damage in order to isolate damage effects to a particular location and investigate how this location may asymmetrically affect the stability of the quadrotor, if at all. Each flight condition - that is, combination of damage level, damage location, and flight tracking task - is summarized in table II. Note that some flight conditions could not be completed reliably as they resulted in frequent flyways (such as Supplementary Video 3), leading to fewer than five successful flights for the condition. In total, 247 individual flights, lasting around one minute each, are completed without a crash occurring.

We rely on the rotor speed RPM (revolutions per minute) measurements, $\boldsymbol{\omega}$, as the basis for our dynamical indicators of quadrotor instability. This is because $\boldsymbol{\omega}$ serves as both the input to the quadrotor system and the (motor dynamics filtered) output of the controller (see Supplementary Information Section VI for context on the quadrotor dynamics). As such, $\boldsymbol{\omega}$ captures the (potentially unstable) interaction between the controller and the system. We measure $\boldsymbol{\omega}$ through the onboard electronic speed controller (ESC) via the Bidirecitonal DShot protocol. This data is logged at 1000 Hz. Since *INDIFlight* applies less filters to the raw sensor data prior to logging than *Betaflight*, we first apply a 100 Hz low-pass fourth order Butterworth filter to limit the power of high-frequency noise in the measurements to approximate the low-pass filtering conducted by *Betaflight* prior to data logging. Subsequently, we downsample the *INDIFlight* measurements to match the 500 Hz logging rate of *Betaflight*. These filtering steps (i.e. 100 Hz low-pass filter and downsampling) result in an *INDIFlight* rotor speed frequency spectra that is similar to *Betaflight*, affording the use of consistent critical slowing down parameters across the two systems.

*3) Human-piloted flight experiments:* The piloted flight tests of the *BetaFlight* controlled HoverFly serve as means to

[i] Freely available here: https://github.com/tudelft/INDIFlight.git

[ii] Freely available here: https://www.betaflight.com/download

compare propeller damage tolerance under different flight control software. Due to a lack of autonomous capabilities, only hovering flights are conducted to promote consistency in the pilot's behavior. Five hovering flights, lasting around one minute each, are flown at each damage level (from 0% to 55% damage as shown in fig. 1 **c**) and each rotor location. As with the *INDIFlight* controlled DragonFly, only one propeller is damaged at a time. Although the hovering flight task is less demanding than the autonomous maneuvers conducted with the DragonFly, manual hovering flight with the DragonFly is already impossible at the 30% damage level due to persistent flyaways (such manual flight tests with *INDIFlight* are shown in Supplementary Video 1). Thus, the improved resilience of the *Betaflight* controlled HoverFly is not due to the less demanding maneuvers alone but is primarily a function of the controller itself. In total, 120 flights of the HoverFly are flown for which the rotor speed RPM measurements are provided by the onboard ESC via Bidirectional DShot, logged at 500 Hz.

### *B. Generic indicators of instability*

Critical slowing down (CSD) is used as an early warning of critical transitions across a family of bifurcations[28,40]. At these bifurcation points, the dominant eigenvalue, $\lambda_{max}$, of the system's Jacobian matrix exchanges stability. Mathematically, the real part of this eigenvalue approaches zero (i.e. $|Re(\lambda_{max})| \rightarrow 0$) near the bifurcation point and, as a consequence, the rate of recovery of the system to equilibrium slows down. It is this change in rate of recovery that CSD is sensitive to. Analogously to bifurcations, as a controlled system (locally) approaches instability, the dominant eigenvalue of the associated closed-loop (i.e. combination of system and feedback controller) Jacobian also approaches zero: $Re(\lambda_{max}) \rightarrow 0$. Therefore, CSD is well-equipped to monitor shifts towards instability of nominally stable (controlled) systems.

To this end, we employ the lag-1 autocorrelation (AC1) metric of CSD as our generic indicator of controlled system instability. The motivation for this is twofold. First, as a correlation metric, the AC1 enjoys a well-defined domain: $AC1 \in [-1,1]$. Second, the increase in AC1 as a critical transition is approached (i.e. $AC1 \rightarrow 1$) is often consistent[49]. These characteristics facilitate a generic comparison between variations of the same controlled system nearing a critical transition. We calculate the AC1 for the quadrotor using each of the rotor speed measurements, $\boldsymbol{\omega}$, since instabilities driven, in part, by the controller are observable through $\boldsymbol{\omega}$ (see quadrotor dynamics in Supplementary Information Section VI).

*1) Continuous instability monitor:* The generic indicators are calculated along a moving window sliding over the flight data timeseries - rather than over non-overlapping segments of flight - to reflect the intended application of our indicators as a continuous monitor of instability. While we do not compute the indicators in real-time in our experiments, we nonetheless demonstrate its sensitivity for such a task via this rolling window approach. Furthermore, we note that incremental damage may be incurred during operation or between missions (e.g. damage during transportation) for which the rolling window approach is well-suited to promptly detect a corresponding shift in stability.

*2) Lag-1 autocorrelation per rotor:* For each of the four rotor speed measurements, we first estimate slow trends in the timeseries through a moving average taken within a sliding window along the timeseries. These slow trends are then subtracted from the original signal to detrend it. We chose a moving average filter for its simplicity and suitability for real-time use to detrend new measurement data as it becomes available. Here, we elect a window size of 3 samples (equivalent to 0.006 seconds for our data) to arrive at a detrended (i.e. residual) signal which resembles a first order autoregressive process (see fig. 4).

Subsequently, we estimate the AC1 via a sliding window running along each of the detrended rotor speed measurements individually. Within this window, the AC1 coefficient is estimated via a Pearson correlation of the detrended signal with itself lagged by one sample. We heuristically select a window size of 300 samples (equal to 0.6 seconds) such that the nominal (i.e. no damage) AC1 values are centered near $AC1 \approx 0.5$. Doing so affords space for shifts in AC1 to be observed.

This objective of detecting shifts in stability due to system variations, rather than detecting the exact moment of tipping itself, motivates our choice to estimate the AC1 via a Pearson correlation instead of the more conventional ordinary least squares estimator. The Pearson correlation ensures that $AC1 \in [-1,1]$, providing a consistent domain mapping datum upon which incremental variations of a (controlled) system can be compared whereas OLS does not provide such guarantees on the bounds of the estimated AC1 coefficient.

An example of how the AC1 timeseries are constructed is illustrated in fig. 4, wherein the columns delineate the process per rotor. The result is a set of four histograms of the AC1 values observed for each rotor during the flight. This figure shows that most of the increase in AC1 is due to the damaged rotor itself, whereas the undamaged rotors experience limited shifts in AC1. Despite this, to reflect real world operations, we observe the shifts among all rotors simultaneously as the location of the damage is often unknown a-priori (though its location may be diagnosed through our indicator). Therefore, to instead visualize the holistic shift in stability of the quadrotor due to damage (as in fig. 2), we concatenate the AC1 values derived from each of the four rotors into a single representative AC1 frequency histogram per damage level (e.g. 10%) and damage location (e.g. front right). That is, each histogram in fig. 2 includes data of all rotor measurements across all the relevant flight tasks (summarized in table II). Note that this in fact disadvantages the sensitivity of our indicator for detecting loss of stability.

*C. Statistical analyses*

*1) Significance testing:* Since the lag-1 autocorrelation (AC1) values of a given flight task are computed over a rolling window, they are (time) dependent. Therefore, for statistical analyses, we collect the median AC1 values of each rotor per flight of the DragonFly (i.e. each flight yields four independent median AC1 values, one from each rotor location). We choose the median as it reflects the center point of the underlying AC1 histogram and is robust against outliers that can occur during flight due to abnormal conditions (e.g. brief WiFi dropouts).

We then aggregate these median AC1 values across damage conditions of interest to form a representative AC1 distribution (see "Damage comparison" in table III for the damage conditions considered in our analyses). As an example, the median AC1 distribution that represents the nominal condition is composed of the median AC1 values of each flight of the DragonFly with 0% damage (first row in table II). This yields an AC1 distribution with 200 data points (50 flights with 4 median values each, representing each rotor measurement).

Subsequently, the significance of increases in these median AC1 distributions due to propeller damage is assessed through a (one-sided) Brunner Munzel nonparametric test. This test is used to evaluate the probability, P, that a random sample from a damaged AC1 distribution of interest is larger than a random sample from the reference nominal AC1 distribution, under the null hypothesis that this probability is equal (i.e. $\mathrm{P} = 0.5$). The null hypothesis implies that the distributions are stochastically the same. Note that we favor the Brunner Munzel test over the Mann-Whitney U test since the former allows for unequal variances between the two distributions, which is the case for our data: the damaged AC1 distributions tend to harbor larger variances than the nominal distribution. These differences arise from nonlinear propeller damage effects between the various maneuvers conducted across the flight experiments. We consider a result significant if the Brunner Munzel p-value is less than 5% (i.e. $p < 0.05$).

A natural effect size measure for the Brunner Munzel test is to evaluate the Probability of Superiority (Vargha-Delaney A), $PS$, which quantifies how likely it is that a random AC1 sample from distribution $A$ (e.g. damaged) is greater than a random AC1 sample from distribution $B$ (e.g. nominal):

$$PS = P(A > B) + \frac{1}{2}P(A = B)$$

In other words, the $PS$ represents the consistency of increases in AC1 due to propeller damage, relative to the nominal case. Here, we assign half credit in the event of ties between the AC1 values of the damaged and undamaged distributions. Thus, if these distributions are indeed the same (i.e. $A = B$ for all values) then $PS = 0.5$.

We likewise measure the corresponding magnitude of increase in AC1 through the Hodges' Lehman shift, scaled by the inter-quartile range of the nominal AC1 distribution to contextualize this effect size. That is, how much has the AC1 increased in comparison to the inter-quartile range of the nominal AC1 baseline.

The statistical analysis results are summarized in table III and show that damage incurred by the propellers significantly increases the AC1 values. The magnitude of increase grows alongside the damage level. Also shown in table III are the statistical tests between damage locations (i.e. front versus aft and left versus right) and the corresponding increase in AC1, revealing significant asymmetries in stability for the DragonFly quadrotor.

*2) Indicator parameter sensitivity analysis:* We assess the sensitivity of the median AC1 statistical results as a function of the generic indicator parameters (i.e. moving average detrender window size and AC1 window size). The moving average window is varied from 3 to 25 samples (0.006 to 0.05 seconds) and AC1 window from 50 to 1500 samples (0.1 to 3.0 seconds). Figure 6 illustrates the change in Brunner Munzel p-value and Probability of Superiority, $PS$, effect size for different combinations of these generic indicator parameters.

The trends in fig. 6 indicate that the increases in AC1 due to damage remain largely significant despite variations in the generic indicator parameters, especially at the 15% damage level. Nevertheless, the Brunner Munzel p-value and $PS$ deteriorate with increasing AC1 window size. In contrast, variations in the moving average window have little effect.

These trends reflect the fact that the quadrotor is an incredibly fast system. Thus, the dynamics of interest often reside at high frequencies (i.e. short time scales), especially in the face of propeller damage. Nonetheless, these high frequency components can bleed into lower frequencies: instabilities become apparent over numerous cycles of the high-frequency effects. Moreover, these statistical results agree with the qualitative observations of instability for the DragonFly quadrotor already at the 15% damage level (see Supplementary Video 3).

## **Box 1** | Effect of delay in feedback systems

Real systems harbor inherent delays. Consider, for example, the delays in blood pressure regulation after one stands up rapidly or the lag in temperature change that follows the adjustment of a thermostat. These delays can induce instability in `closed-loop' systems (i.e. controlled systems for which the controller's actions depend on the system outputs). We illustrate this using a simple second order (linear) system (eq. B-1), though the destabilizing effect of input-output delay extends to real-world nonlinear systems. Equation B-1 is expressed in the Laplace domain with *s* the complex Laplace variable, $\omega_n$ denotes the natural frequency of the system and $\zeta$ is its damping ratio. Let $\omega_n = 1$ and $\zeta = 0.5$.

$$H(s) = \frac{\omega_n^2}{s^2 + 2\zeta\omega_n s + \omega_n^2} \qquad \text{(B-1)}$$

Box 1 figure **a** illustrates the inherent delay of $P = H(s)$ in `open-loop' (i.e. inputs, dashed line, made by the controller are invariant of the system output, solid line). If this delay is known, it can be compensated for in the control design to achieve the desired tracking behavior. For example, we design eq. B-2 for near perfect tracking of the input signal, seen in Box 1 figure **b** with $P = H(s)$.

$$C(s) = 4.5 + \frac{0.5}{s} \qquad \text{(B-2)}$$

However, additional delays can arise in real-world systems, for instance, due to actuator dynamics and faults, time-varying dynamics, natural component degradation or even environmental conditions that may limit the effectiveness of the controller.

In this example, we consider the presence of first order actuator dynamics (eq. B-3) which increase the delay of the open-loop system as shown in Box 1 figure **c** with $P = H_{act}(s)H(s)$.

$$H_{act}(s) = \frac{1}{2s + 1} \qquad \text{(B-3)}$$

Then, under the same control law (eq. B-2) the closed-loop system now becomes unstable, depicted in Box 1 figure **d** with $P = H_{act}(s)H(s)$. This is because a positive (i.e. self-reinforcing) feedback loop in the input-output relation arises; the controller attempts to regulate the system's response but only ends up amplifying it.

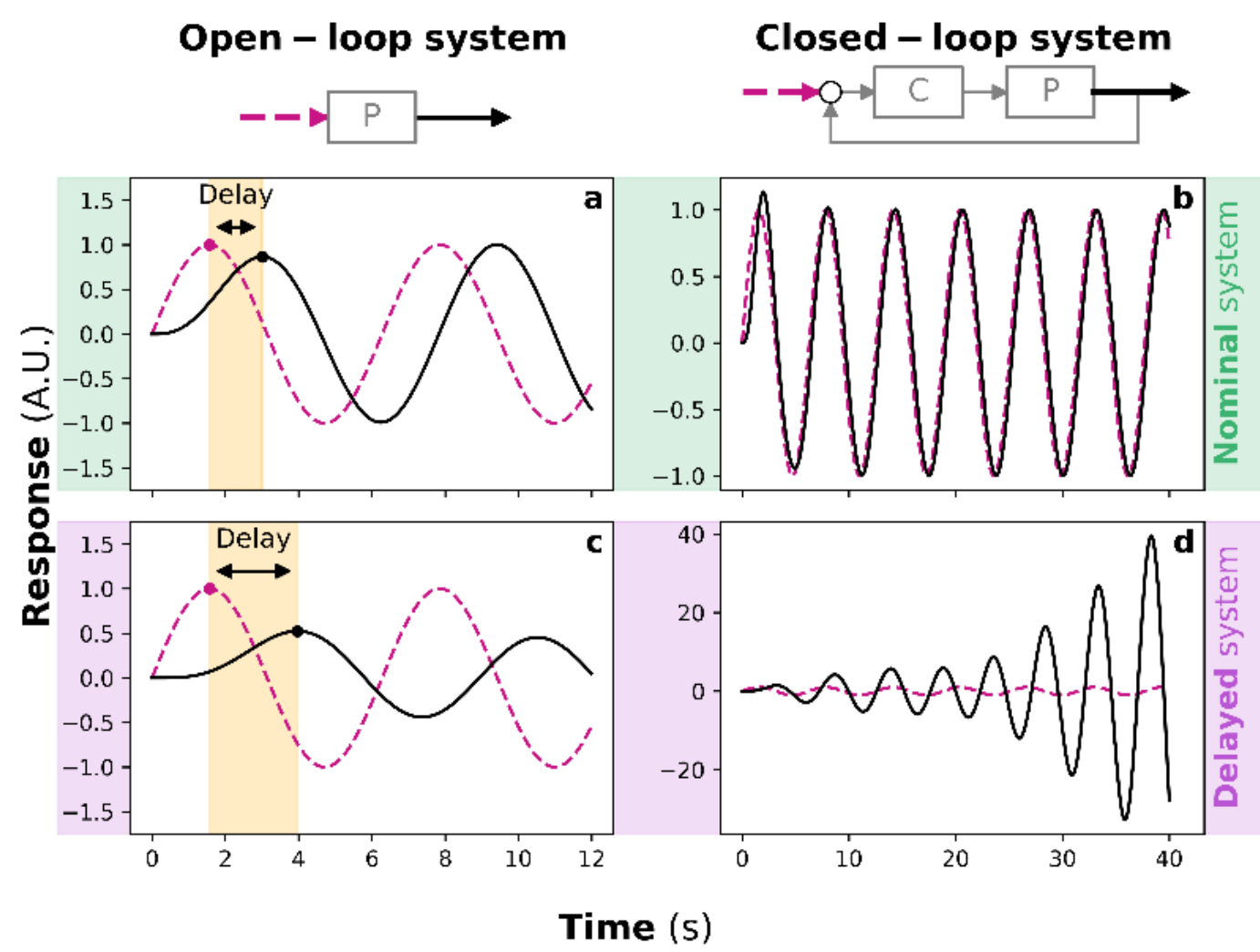

**Box 2** | State measurement imperfections

Real-world feedback controllers rely on sensor measurements of the system states for control. These sensors are often imperfect and contaminate the measurement signal with artifacts that the controller should ignore, such as biases, noise, additional (sensor) dynamics and sensitivities. Fortunately, these sensor imperfections can be mitigated through clever and targeted filtering. This equips the controller with authority around the frequency ranges of interest and no (or limited) authority around problematic frequency regions (e.g. resonance). While these filters work as desired for the nominal system, issues can re-emerge once the system dynamics change (e.g. due to faults). Similarly, additional delays incurred by the system - which alone are perhaps unproblematic - can destabilize the system when used in tandem with the nominal filters.

We illustrate such a destabilizing effect using the same controlled system as Box 1 (i.e. eq B-1 and eq. B-2) which is now driven by a faster first order actuator (eq. B-4) and relies on a sensor (eq. B-5) for state feedback.

$$H_{act}(s) = \frac{1}{0.1s + 1} \tag{B-4}$$

$$H_{sens}(s) = \frac{4.5^2}{s^2 + 1.35s + 4.5^2} \tag{B-5}$$

The individual effects of input-output delay and sensor resonance do not result in instability (Box 2 figure **b** and **c** respectively). However, their combined effect does (Box 2 figure **d**).

Note that similar resonant sensor dynamics of form eq. B-5 are common of many micro-electro-mechanical systems (MEMS), such as those typically found on a quadrotor[50,51,52]. These resonant modes can get excited by vibrations from propeller damage, deteriorating the quality of the feedback signal.

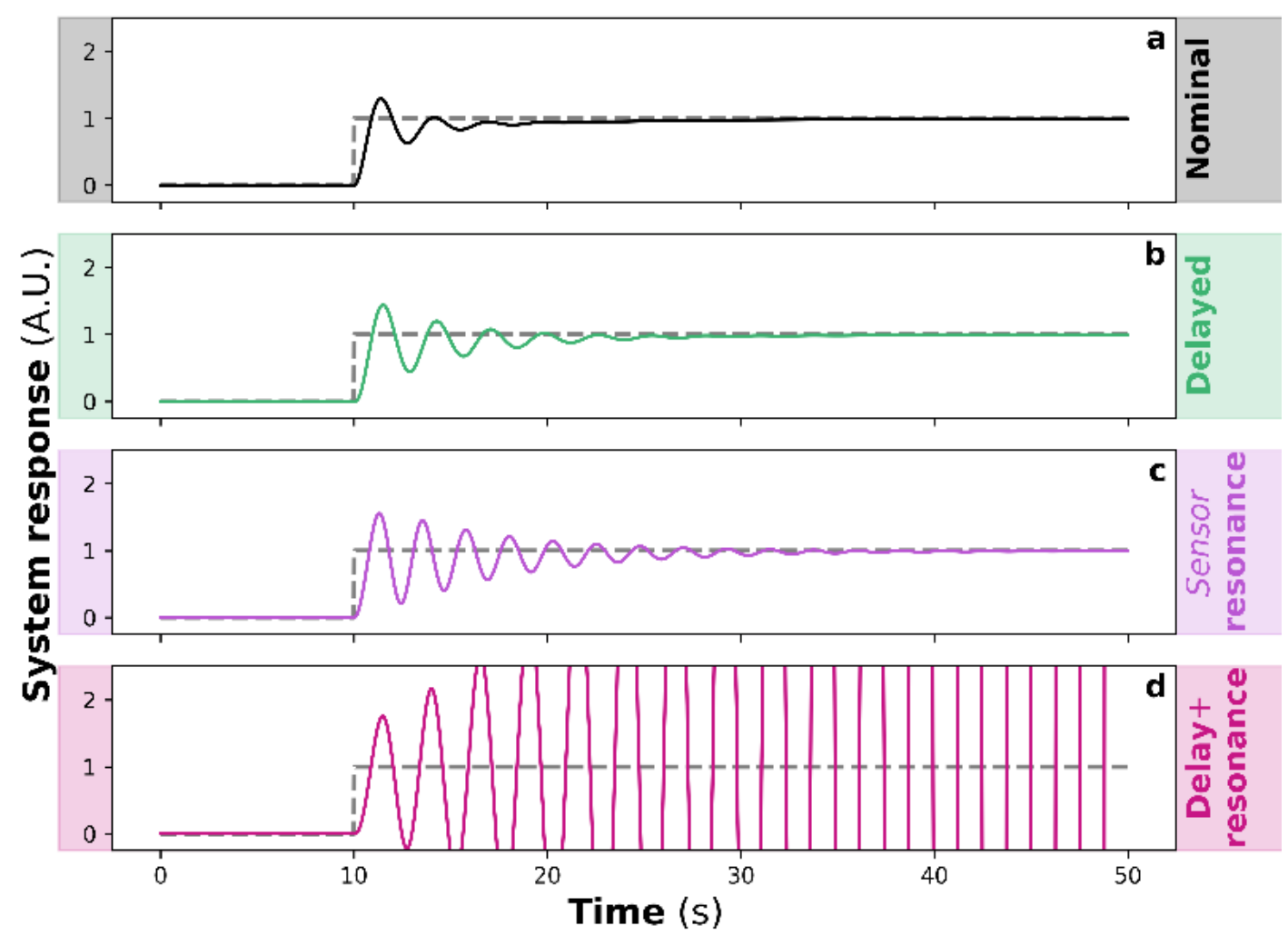

## Method References

## Acknowledgments

This work is supported through The Dutch Research Council (NWO) VIDI grant 18378 on “forecasting safe operating envelopes for autonomous robots”.

## Author contributions

J.J.B. formulated the main ideas, designed and conducted the flight experiments, performed the data & statistical analysis, created the display items, and wrote the paper. M.S. formulated the main ideas, the experiment design, contributed to data analysis and writing the paper. P.S. contributed to the main ideas, experiment design, aided with data collection, analysis of results, and writing the paper. I.A.L. contributed to the main ideas, statistical & data analysis, and writing the paper. E.H.N. contributed to the main ideas, experiment design, statistical & data analysis and interpretation, and writing the paper. C.C.V. formulated the main ideas, the experiment design, contributed to analysis & interpretation of results, writing the paper, and provided funding.

# V. Extended data

TABLE I

Quadrotor properties. Components and physical characteristics of the two quadrotors used in the flight experiments.

| | **DragonFly** (fig. 1 **a**) | **HoverFly** (fig. 1 **b**) |
|---|---|---|
| **Drone frame** | Ethix CineRat | Ethix CineRat |
| **Mass (incl. batteries),** $g$ | 494.13 | 282.56 |
| **Diagonal hub-to-hub diameter,** $mm$ | 153 | 153 |
| **Propeller diameter,** $mm$ | 76 | 76 |
| **Motor** | Emax Eco 1407 2800 kV | Ethix Flat Rat-V2 1507 2800 kV |
| **Batteries** | Tattu FunFly 22.2V 1300mAh | Tattu R-Line 14.8V 850mAh |
| **Flight Controller (FC)** | MATEKSYS H743-SLIM V3 | MATEKSYS H743-SLIM V3 |
| **FC Software** | INDIFlight | Betaflight 4.3.2 |

TABLE II

Autonomous flight conditions. Overview of autonomous (indiflight) flight tasks of the DragonFly.

| **Damage location** | **Damage level** | **Flight task** (see fig. 3) | **Windy** | **Number of flights** |
|---|---|---|---|---|
| N/A | 0% | Hover | No | 10 |
| 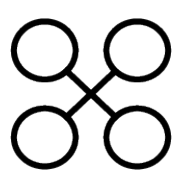 | 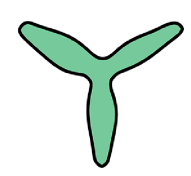 | Circle at 1.5m/s | No | 10 |
| | | Circle at 1.5m/s | Yes | 5 |
| | | Circle at 3.0m/s | No | 5 |
| | | Rectangle (small) | No | 10 |
| | | Rectangle (small) | Yes | 5 |
| | | Rectangle (large) | No | 5 |
| Front right | 10% | Hover | No | 10 |
| 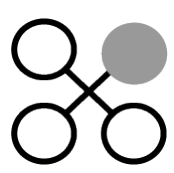 | 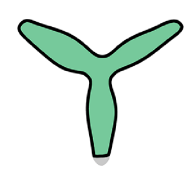 | Circle at 1.5m/s | No | 5 |
| | | Circle at 1.5m/s | Yes | 5 |
| | | Circle at 3.0m/s | No | 5 |
| | | Rectangle (small) | No | 5 |
| | | Rectangle (large) | No | 5 |
| Front right | 15% | Hover | No | 10 |
| 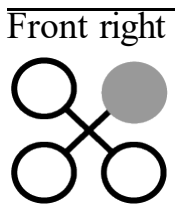 | 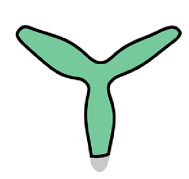 | Circle at 1.5m/s | No | 5 |
| | | Circle at 1.5m/s | Yes | 5 |
| | | Circle at 3.0m/s | No | 4 |
| | | Rectangle (small) | No | 5 |
| Aft right | 10% | Hover | No | 10 |
| 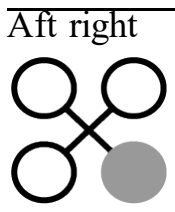 | 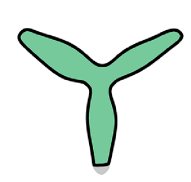 | Circle at 1.5m/s | No | 5 |
| | | Circle at 1.5m/s | Yes | 5 |
| | | Circle at 3.0m/s | No | 5 |
| | | Rectangle (small) | No | 5 |
| Aft right | 15% | Hover | No | 10 |
| 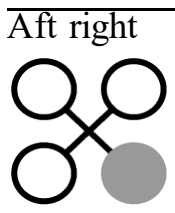 | 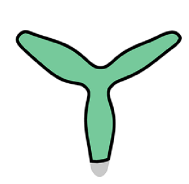 | Circle at 1.5m/s | No | 5 |
| | | Circle at 1.5m/s | Yes | 5 |
| | | Circle at 3.0m/s | No | 3 |
| | | Rectangle (small) | No | 5 |
| Front left | 10% | Hover | No | 5 |
| 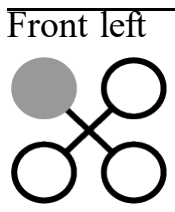 | 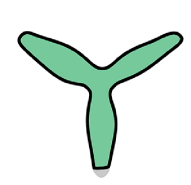 | Circle at 1.5m/s | No | 5 |
| | | Rectangle (small) | No | 5 |
| | | Rectangle (small) | Yes | 5 |
| Front left | 15% | Hover | No | 5 |
| 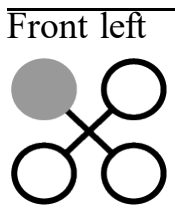 | 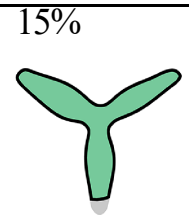 | Circle at 1.5m/s | No | 5 |
| | | Rectangle (small) | No | 5 |
| | | Rectangle (small) | Yes | 5 |
| Aft left | 10% | Hover | No | 5 |
| 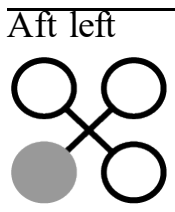 | 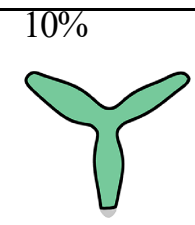 | Circle at 1.5m/s | No | 5 |
| | | Rectangle (small) | No | 5 |
| | | Rectangle (small) | Yes | 5 |
| Aft left | 15% | Hover | No | 5 |
| 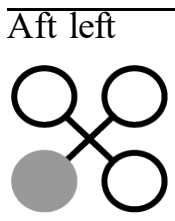 | 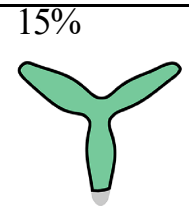 | Rectangle (small) | No | 5 |
| | | Rectangle (small) | Yes | 5 |

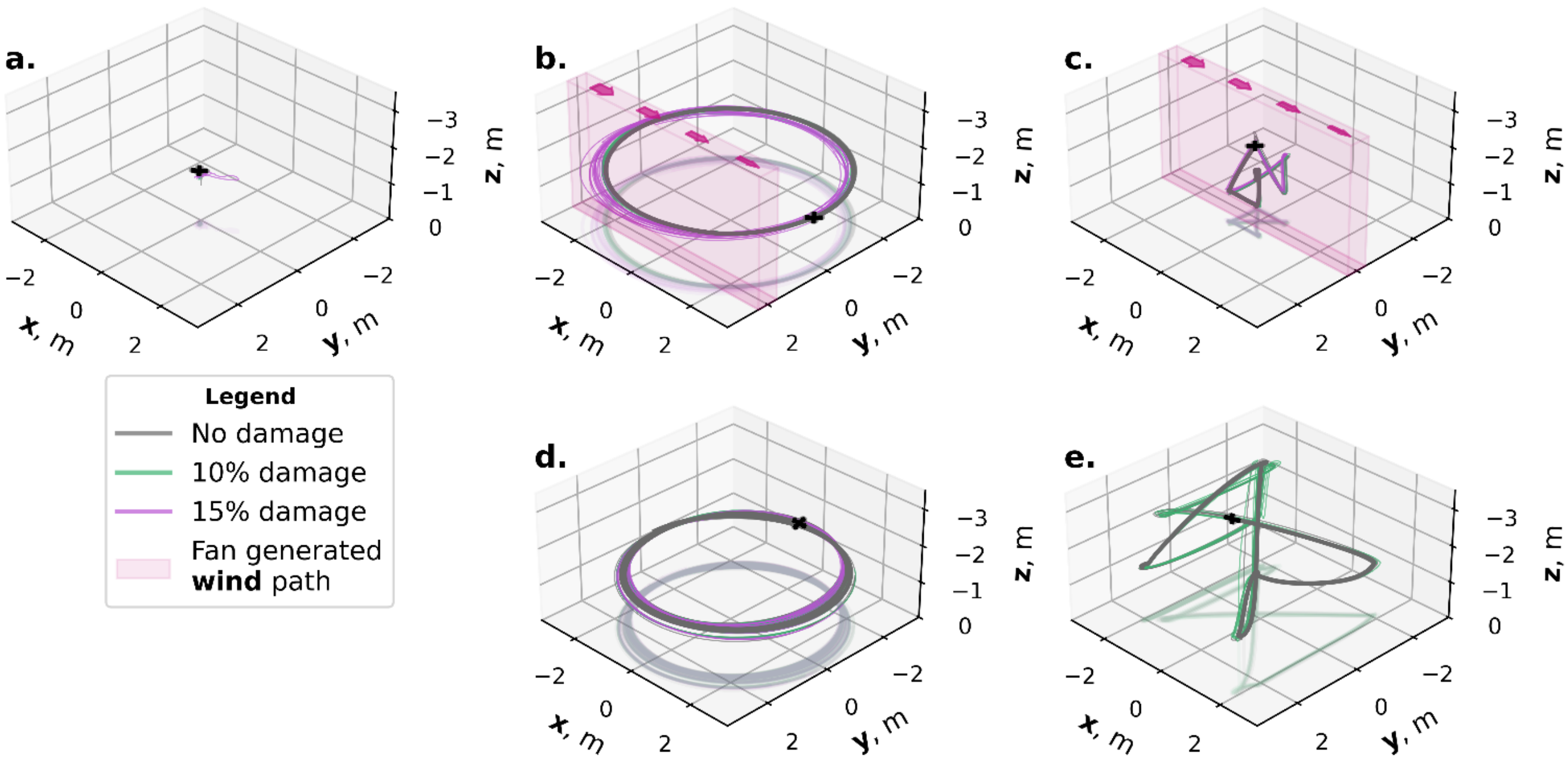

**Figure 3 | Autonomous flight trajectories.** Illustration of the various flight trajectories flown during the experimental campaign of the *INDIFlight* controlled DragonFly. **a** depicts a set of trajectories of the hovering flight task at each damage level. Examples of circular flight tasks are shown in **b** and **d** at tracking speeds of 1.5 and 3.0 m/s respectively. Likewise, **c** and **e** illustrate the trajectories of the ‘small’ and ‘large’ rectangle vertex set-point tracking tasks, respectively. Moreover, **b** and **c** involve both windless and windy flights, wherein the region of influence and direction of the wind (when active) is highlighted by the shaded region. In all plots, the colors denote the damage level, and the opaque lines depict the projection of these trajectories on the x-y plane for spatial context.

TABLE III

Statistical analyses (median). Statistical analyses for increases in lag-1 autocorrelation (AC1) medians due the presence of rotor damage.*

| Damage comparison | Num. $A$, Num. $B$ | PS, P($A$ > $B$) [95% CI]† | HL shift [95% CI]‡ | HL / IQR($B$) | BM p-value § |
|---|---|---|---|---|---|
| **10% dmg** ($A$) **> No dmg** ($B$) | | | | | |
| All rotors | 420, 200 | 0.59 [0.54, 0.63] | 6.1e-3 [2.8e-3, 9.6e-3] | 0.26 | < 0.001 |
| Front rotors only | 220, 200 | 0.54 [0.49, 0.60] | 2.9e-3 [-8.0e-4, 6.6e-3] | 0.12 | 0.064 |
| Aft rotors only | 200, 200 | 0.64 [0.58, 0.69] | 1.0e-2 [5.9e-3, 1.5e-2] | 0.43 | < 0.001 |
| Right side rotors only | 260, 200 | 0.57 [0.52, 0.62] | 4.8e-3 [1.2e-3, 8.7e-3] | 0.21 | 0.005 |
| Left side rotors only | 160, 200 | 0.62 [0.56, 0.68] | 8.2e-3 [4.0e-3, 1.2e-2] | 0.35 | < 0.001 |
| **15% dmg** ($A$) **> No dmg** ($B$) | | | | | |
| All rotors | 368, 200 | 0.76 [0.72, 0.80] | 2.7e-2 [2.1e-2, 3.3e-2] | 1.14 | < 0.001 |
| Front rotors only | 196, 200 | 0.70 [0.65, 0.75] | 2.1e-2 [1.5e-2, 3.0e-2] | 0.91 | < 0.001 |
| Aft rotors only | 172, 200 | 0.83 [0.79, 0.87] | 3.1e-2 [2.5e-2, 3.6e-2] | 1.31 | < 0.001 |
| Right side rotors only | 228, 200 | 0.76 [0.71, 0.81] | 2.3e-2 [1.8e-2, 2.8e-2] | 0.98 | < 0.001 |
| Left side motors only | 140, 200 | 0.76 [0.71, 0.82] | 3.5e-2 [2.7e-2, 4.4e-2] | 1.51 | < 0.001 |
| **Aft** ($A$) **> Front** ($B$) | | | | | |
| 10% damage | 200, 220 | 0.60 [0.54, 0.65] | 8.5e-3 [3.7e-3, 1.4e-2] | 0.33 | < 0.001 |
| 15% damage | 172, 196 | 0.57 [0.51, 0.63] | 9.5e-3 [1.8e-3, 1.7e-2] | 0.12 | 0.010 |
| **Left** ($A$) **> Right** ($B$) | | | | | |
| 10% damage | 160, 260 | 0.54 [0.49, 0.60] | 3.6e-3 [-1.1e-3, 8.4e-3] | 0.12 | 0.066 |
| 15% damage | 140, 228 | 0.55 [0.49, 0.62] | 8.6e-3 [-1.0e-3, 1.9e-2] | 0.17 | 0.044 |

*PS = Probability of superiority, CI = Confidence interval, HL = Hodge’s Lehman shift, IQR(B) = Inter-quartile range of $B$, and BM = Brunner-Munzel. The median AC1 distributions being compared are denoted by $A$ and $B$ respectively, where $B$ acts as the ‘control’ for the BM statistical test.

†PS computed using half credit for ties (i.e. where rank $A$ = rank $B$). The associated CI are obtained via Wald’s CI.

‡The HL shift CI obtained through an asymptotic approximation to the U-statistic.

§The BM test is conducted with the one-sided alternative that $A > B$.

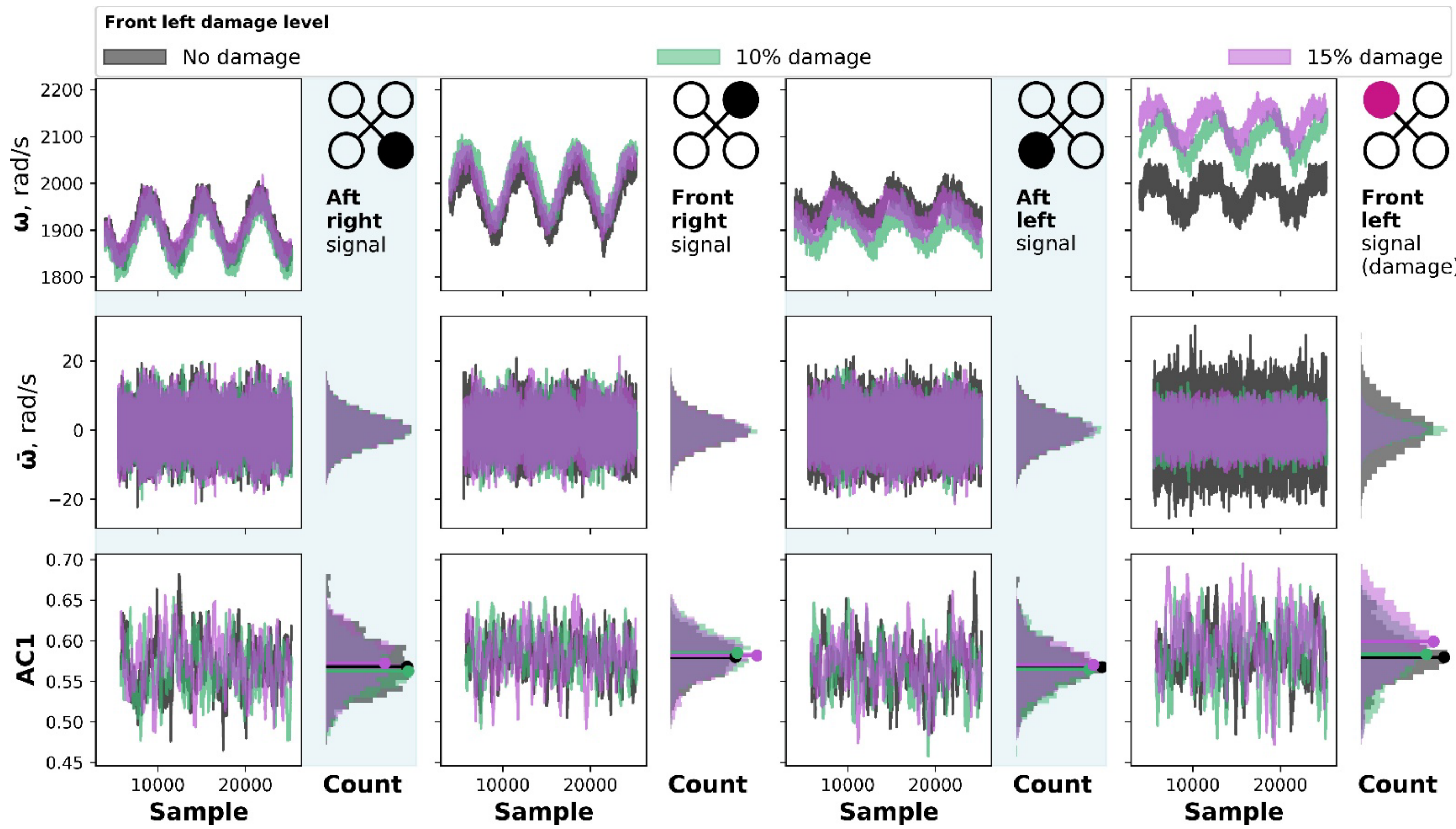

**Figure 4 | Construction of the generic indicators of quadrotor instability.** The generic indicators of quadrotor instability are derived from rotor speed measurements, $\omega$, (top row of plots) obtained from the DragonFly quadrotor's onboard electronic speed controller. Slow trends in these measurements are removed using a moving average detrender (window size of 3 samples, or 0.006 seconds), shown in the middle row of plots. This transforms the detrended rotor speeds, $\overline{\omega}$, into an autoregressive process upon which the lag-1 autocorrelation metric (AC1) of critical slowing down may be applied. The AC1 is calculated over a moving window of 300 samples (equal to 0.6 seconds) sliding over each of the detrended rotor speed signals individually. The resultant AC1 are depicted in the plots along the bottom row. The median AC1 values are highlighted in the accompanying count histograms. Data for these plots comes from three individual windless circular tracking flights of the DragonFly at a speed of 1.5 m/s: (i) no damage to the front left rotor, (ii) 10% damage to the front left rotor, and (iii) 15% damage to the front left rotor. Measurements from all rotor locations (including undamaged rotors) are used in to inform the generic indicators. The columns correspond to the different rotor locations where the solid circle indicates the propeller location from which the measurement is obtained. The damaged rotor (front left) is denoted by the pink circle in the rightmost column.

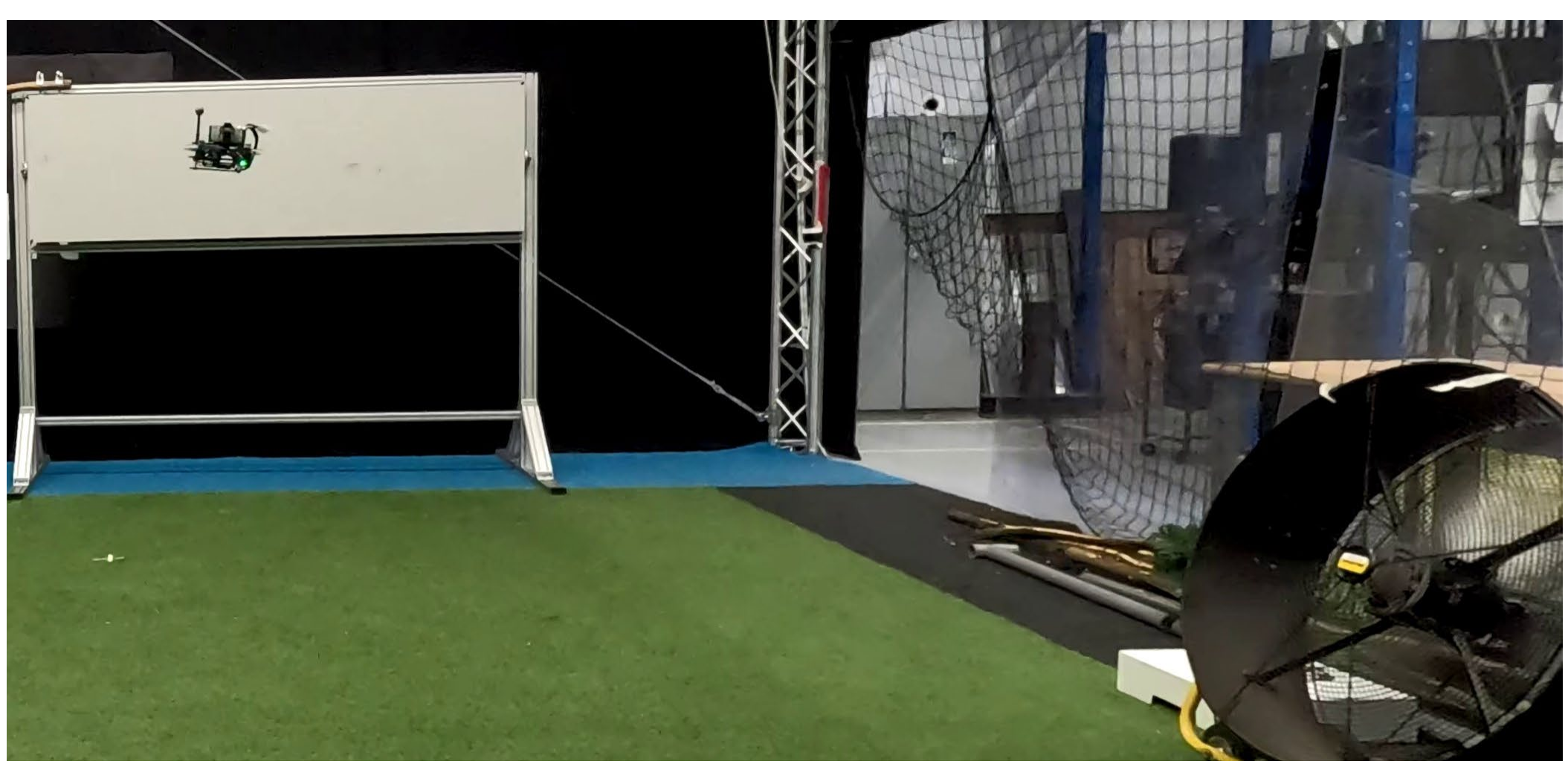

**Figure 5 | Windy flight experiments.** Snapshot of one of the autonomous flight experiments with wind generated by a large fan.

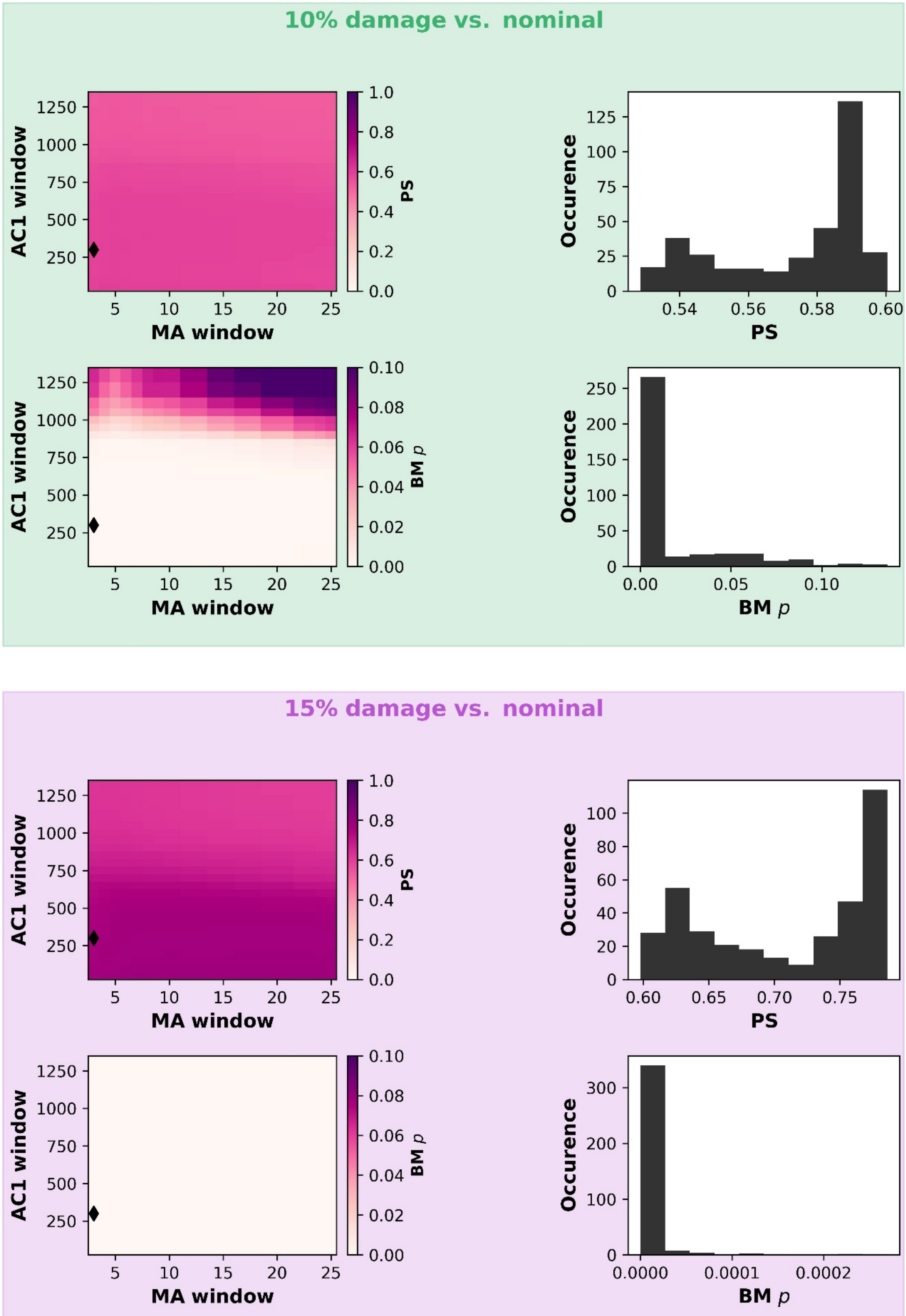

**Figure 6 | Sensitivity of statistical results (median).** The generic indicators of instability are calculated over a moving average (MA) detrender window and lag-1 autocorrelation (AC1) window. The sensitivity of the significance of increases in AC1 (due to 10% and 15% propeller damage) with respect to variations in these parameters are depicted in the heatmaps along the left column plots, organized by damage degree. The right column plots provide histograms of the associated Brunner Munzel (BM) statistical test p-values and probability of superiority, PS, effect sizes.

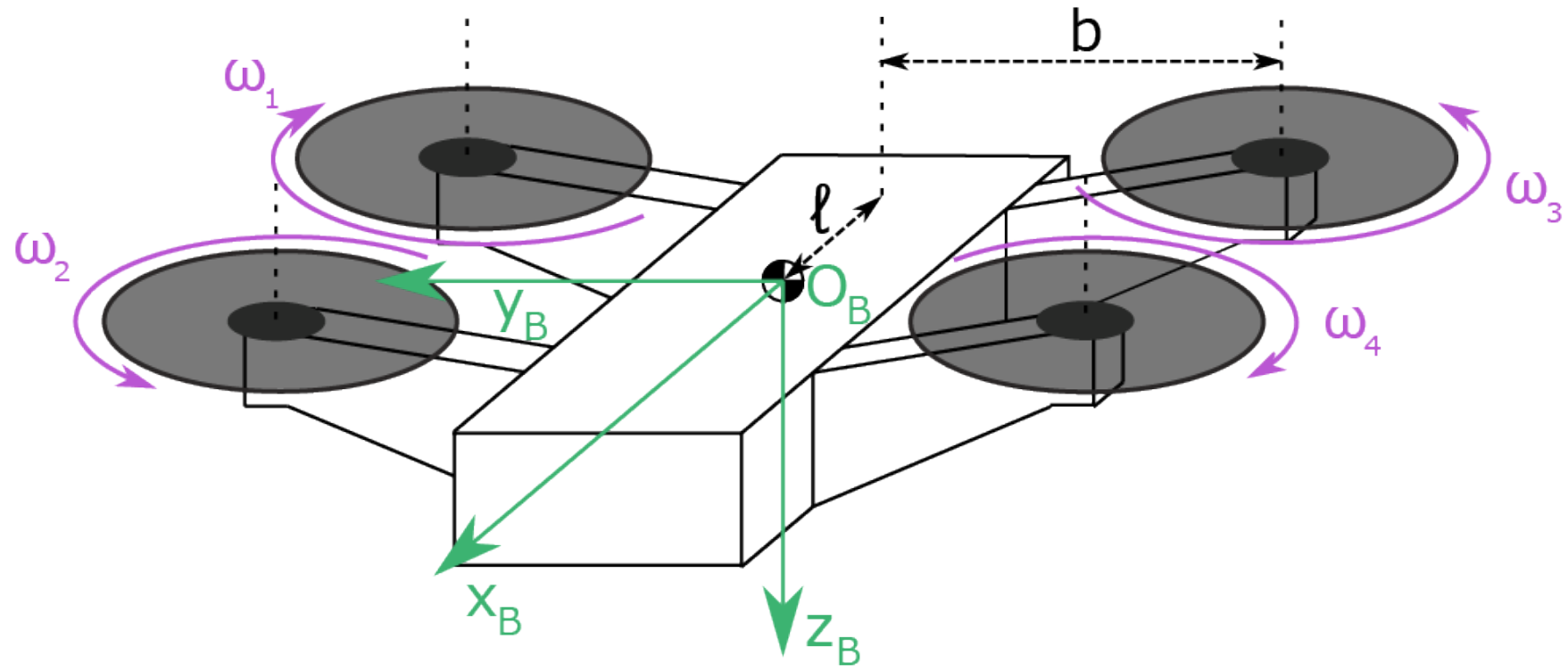

**Figure 7** | **Coordinate frame and geometric properties of the quadrotor.** The rotor layout utilized by both our quadrotors is indicated by $\omega \in \{\omega_1, \omega_2, \omega_3, \omega_4\}$ in purple with rotor 1 ($\omega_1$) spinning clockwise. The forward-right-down body reference frame is shown in $O_B = \{x_B, y_B, z_B\}$. $b$ and $\ell$ describe the x and y moment arms, respectively.

## VI. Supplementary information: Quadrotor system model

While our generic indicators of instability do not depend on a system model, we nonetheless provide the kinematic equations of the quadrotor here to: (i) motivate the choice of the rotor speeds as the early warning variables, and (ii) provide context on the nonlinear dynamics underlying the quadrotor. The quadrotor rigid body dynamics can be written as:

$$\begin{bmatrix} \dot{\boldsymbol{V}} \\ \dot{\boldsymbol{\Omega}} \end{bmatrix} = \begin{bmatrix} R\boldsymbol{g} - \boldsymbol{\Omega} \times \boldsymbol{V} + \frac{1}{m}\left(\boldsymbol{F} - \kappa \sum_{i=1}^{4} \omega_i^2 \boldsymbol{z_B}\right) \\ I_v^{-1}(-\boldsymbol{\Omega} \times I_v \boldsymbol{\Omega} + \mathbf{M} + \mathrm{J}(\boldsymbol{\omega} \cdot \boldsymbol{\omega}^T)) \end{bmatrix}$$

where $\mathbf{V}$ denotes the quadrotor's translational velocity and $\boldsymbol{\Omega}$ describes its rotational velocity, both expressed in the body front-right-down coordinate reference frame (see fig. 7). The transformation between this body frame and the inertial north-east-down (NED) reference frame is captured by rotation matrix $R$. $\boldsymbol{g}$ represents the gravity vector in the NED frame. The quadrotor mass and vehicle moment of inertia are denoted by $m$ and $I_v$ respectively. The thrust produced by the rotation of the propellers acts in the negative $\boldsymbol{z_B}$ direction (see fig. 7) and is described by $\kappa \sum_i^4 \omega_i^2$, where $\omega_i$ is the rotational speed of the $i^{\text{th}}$ propeller and $\kappa$ is the thrust coefficient. Likewise, the torques produced by the rotor speeds, $\boldsymbol{\omega} = [\omega_1, \omega_2, \omega_3, \omega_4]^T$, is given by $J(\boldsymbol{\omega} \cdot \boldsymbol{\omega}^T)$ with $J$ the (torque) control allocation matrix. Finally, $\boldsymbol{F}$ and $\boldsymbol{M}$ describe any additional (nonlinear) forces and moments acting on the quadrotor body, for example, due to rotor interaction effects or the presence of attached loads. Indeed, the nonlinear dynamics underlying $\boldsymbol{F}$ and $\boldsymbol{M}$ are highly system and application dependent. Nonetheless, our generic indicators of instability may be applied as they are not model dependent.

To this end, we base these indicators on the rotational speed measurements of the propellers (i.e. $\boldsymbol{\omega}$). This is because $\boldsymbol{\omega}$ serves as the (general) input to the quadrotor system and describes the (motor dynamics filtered) output of the controller.

## VII. Supplementary information: Relation to control theory

In this section, we show that critical slowing down (CSD) naturally aligns with control theoretic notions of robustness and reachability. In particular, a decrease in resilience as measured by dynamical indicators of CSD corresponds to a decline in robustness and shrinkage in the backward reachable set. This evidence supports the use of CSD as a generic (i.e. system model-free) indicator to complement traditional means of assessing control system robustness.

To illustrate this link more concretely, consider the following forced mass spring damper system:

$$m\ddot{x} + d\dot{x} + kx = u \tag{1}$$

where $m$, $d$, and $k$ are positive constants (see section VII-E), $x$ denotes the position and $u$ the forcing input, provided by an actuator. An actuator is used here to introduce realistic delay into the system. Such delay makes this controlled system unstable if the controller becomes too aggressive (through self-reinforcing feedback back loops as described in box 1). We adopt first order actuator dynamics for simplicity:

$$\dot{u} = -\frac{1}{\tau}u + \frac{1}{\tau}\tilde{u} \tag{2}$$

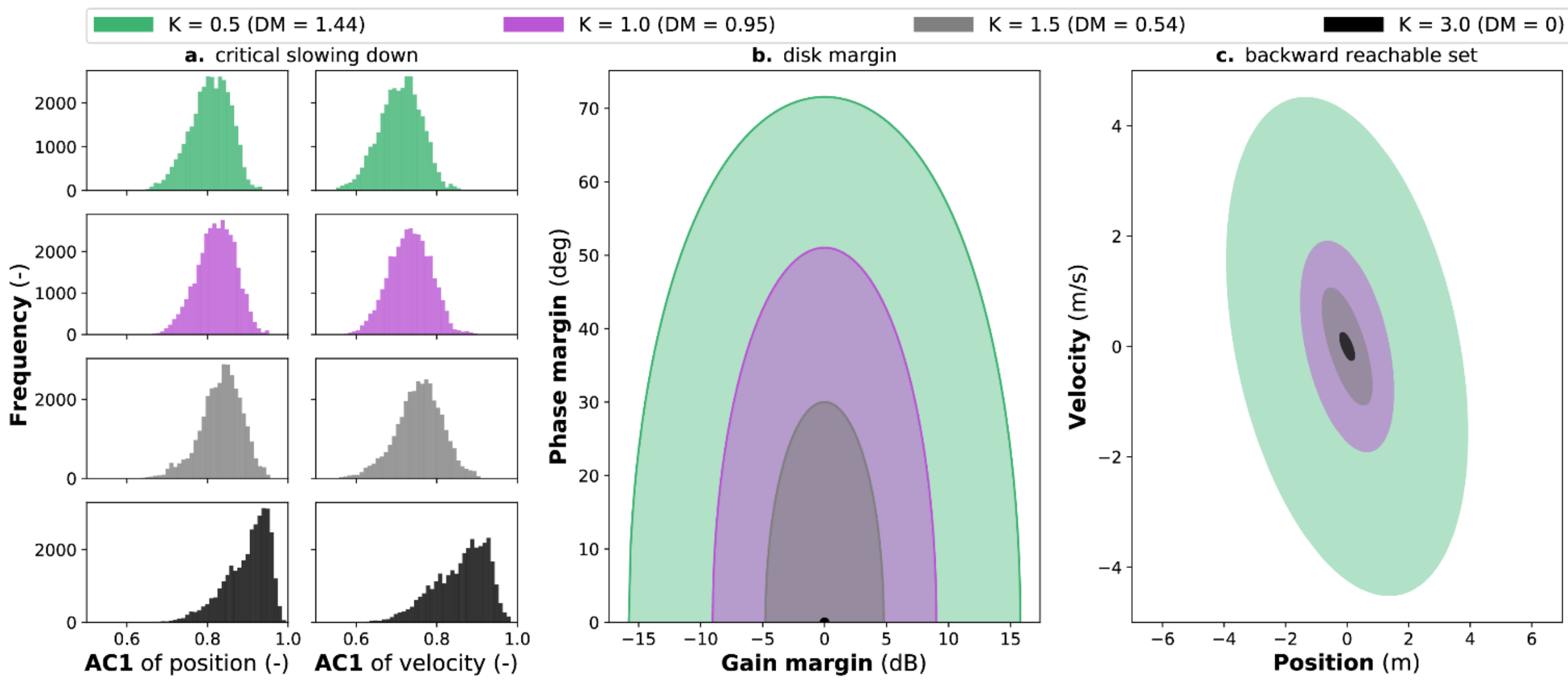

**Figure 8 | Indicators of controlled system resilience**. An actuated (linear) mass spring damper system is used to compare various indicators of controlled system resilience subject to different control gains, K, which govern the stability of the system. **a** depicts the critical slowing down lag-1 autocorrelation (AC1) metric, which shows the characteristic approach to one as the instability is approached incrementally. **b** and **c** are classical control theoretic measures of system robustness and denote the disk margin (DM) and backward reachable set (with an arbitrary time horizon of 12 seconds) respectively. Both exhibit a shrinkage in area as the instability is approached, indicating a loss of robustness (i.e. resilience).

In eq. (2), $\tau$ denotes the actuator rate constant (i.e. how fast the actuator is) and $\tilde{u}$ represents the input commanded by the controller:

$$\tilde{u} = -Kx \tag{3}$$

with $K > 0$ a constant denoting the controller gain (i.e. how aggressive the controller's response to state changes is). Thus, the objective of this controller is to regulate the position of the system to zero (i.e. the equilibrium of the system).

### *A. Generic indicators of instability*

The feedback gain, $K$, commanding the actuated mass spring damper system can be considered a critical parameter since its value governs the stability of the controlled (i.e. closed-loop) system. We note that while linear systems do not experience bifurcations (in the sense that there is no 'branching' of equilibria[53]), they can nonetheless exhibit symptoms of critical slowing down (CSD): as a controlled system approaches instability, the real part of at least one (locally linearized) eigenvalue approaches zero. This is the characteristic phenomenon that results in the slowing recovery rate of a nonlinear system experiencing CSD as a bifurcation is approached.

For the chosen values of $m$, $k$, $d$, and $\tau$ in section VII-E, the controlled system becomes unstable when $K > 4.59$ as the real part of the system's dominant eigenvalues become positive. We select four discrete gain values on approach to this instability: $K = \{0.5, 1.0, 1.5, 3.0\}$. Then, for each gain value, the behavior of the controlled system is simulated with stochastic perturbations (details on these simulations can be found in section VII-E2). For each simulation, we calculate the lag-1 autocorrelation (AC1) metric of CSD using a sliding window along each state variable measurement. Further details on how the AC1 values are calculated are outlined in section VII-F.

The characteristic increase in the AC1 as the controlled system approaches instability due to the incremental amplification of the feedback gain, $K$, is visible in fig. 8 **a**. This step-wise increase in AC1 mirrors that of a well-studied ecological system with alternative stable states (eq. (9)) that also experiences CSD[54] as it approaches a bifurcation through increments of its critical parameter (see fig. 9). These results, taken alongside those of the quadrotor, show that controlled systems can exhibit behavior consistent with CSD and generic indicators of resilience are capable of detecting approaching instabilities in such systems.

### *B. Quantifying resilience to perturbations*

Through the lens of control theory, the notion of a system's proximity to instability is often referred to as system 'robustness'.

One such tool to quantify robustness in the linear setting is via the so-called 'disk margin' (DM)[55], which assesses how much variation in the closed-loop system's gain and phase (i.e. magnitude and delay, respectively) can be tolerated before it becomes unstable. In other words, the disk margin evaluates how much *perturbation* is needed to destabilize the system. These perturbations may stem from (uncertainties in) either the controller or the system dynamics itself, such as signal attenuation, amplification, or delays. A larger disk margin indicates that the system can endure greater perturbations and is thereby more robust (i.e. resilient).

We compute the disk margin for the actuated mass spring damper system subject to the incremental amplification of feedback gains ($K \in \{0.5, 1.0, 1.5, 3.0\}$) towards instability. Indeed, the disk margin decreases as the instability is approached (see **b** in fig. 8). Crucially, these results agree with the trends reported by the generic indicators; the AC1 values increase as the gain is amplified. In fact, the disk margin reveals that the closed-loop system can already suffer instabilities at a gain of $K = 3$ (DM = 0) if there is *any* uncertainty. At this gain value, any additional delays in the system (e.g. if the actuator is even marginally slower than expected) will destabilize it. While the disk margin is undoubtedly a powerful analysis tool in this regard, it relies on a linear system model which limits its practicality for real (nonlinear) systems.

To this end, our results suggest that the generic AC1 may be used as a complement to the disk margin in practice. For instance, the AC1 can monitor for shifts in stability on the real-world system in real-time where loss in stability may arise due to, for example, inaccurate or changing system dynamics. A detection of instability, should it occur, would imply a change in (local linear model) disk margin and thus prompts caution and further investigation (e.g. local model re-identification and disk margin analysis).

### C. *Quantifying basins of attraction*

A general perspective on safety that is compatible with nonlinear systems is the notion of the 'backward reachable set' (BRS)[21,56]. These sets are typically constructed with respect to some (desired) target set of states and, if we take the equilibrium (i.e. $x = \dot{x} = 0$) as such a target set, then the BRS describes the initial states from which system trajectories can start at in order to arrive at the equilibrium within an allocated finite time horizon, $T$. While this is not a direct quantification of system instability, a shrinkage of the BRS can occur as an instability is approached since the system becomes less capable of maintaining the desired equilibrium (i.e. the safe operating region contracts). It is therefore frequently used in the context of system safety to establish a set of safe operating states[57]. Intuitively, the larger the BRS, the more resilient the system is as it is can recover from more states. From the perspective of CSD, the system has a larger *basin of attraction*.

Another common approach towards determining the basin of attraction of a nonlinear control system is through Lyapunov stability analysis. Briefly, the goal is to determine whether the trajectories of a controlled system either remain near an equilibrium or approach it asymptotically as time approaches infinity. This relates naturally to the role of the rate of recovery dynamics in CSD: before the bifurcation point, the non-zero rate of recovery ensures that the system's trajectory converges to the equilibrium point at some point in time. However, at the bifurcation point, the rate of recovery vanishes implying that a state trajectory is no longer guaranteed to return to, or even remain near, the equilibrium; the system is no longer stable in the Lyapunov sense. Though Lyapunov stability theory is a useful tool for assessing the time-invariant (asymptotic) stability of a nonlinear control system, we use the BRS here to make explicit the link between CSD and the reachability of the mass spring damper system in a time-varying setting. Moreover, as we are concerned with the changing rate of recovery of a system, we favor approaches that explicitly involve a finite temporal element.

To this end, we compute the BRS for the actuated mass spring damper system, under the different feedback gain values, for a time horizon of 12 seconds (which affords enough time for the system to reach its steady states for the given system constants in section VII-E). The resultant BRS are depicted in **c** of fig. 8 for which a clear shrinkage in the BRS can be observed as the gain, $K$, is increased towards instability. This shrinkage mirrors that of the disk margin and the trends in the generic indicators of CSD (i.e. AC1). Indeed, CSD is known to reflect a change in a system's basin of attraction[28,33] and is therefore a rather intuitive, yet generic, proxy of the BRS.

To further illustrate this connection between the BRS and CSD, we compute the BRS as a function of the critical parameter for the actuated mass spring damper system and a nonlinear ecological system. Specifically, we employ the resource exploitation model (eq. (9)) of Dakos et al.[54] to simulate an ecological system with two alternative stable states. Figure 10 depicts the resultant BRS of the engineered and ecological systems as a function of their critical parameters for various time horizons. The contracting basins of attraction as the respective critical transitions are approached are clearly visible. How the BRS are computed for these systems, alongside the parameters of the over-grazing model, are summarized in section VII-G.

### D. *Critical slowing down and control theory*

These results demonstrate that critical slowing down relates well to classical notions of robustness in control systems. All these frameworks provide different perspectives on system resilience. While the disk margin and backward reachable set can be used to respectively quantify the tolerable perturbations and basin of attraction of a system, they rely on substantial and accurate knowledge of the system model in order accomplish this. Instead, critical slowing down allows us to monitor such changes system characteristics in a generic (i.e. model-free) fashion, at the cost of a lack of a quantification of the exact bounds of attraction or perturbation scale.

### *E. Actuated mass spring damper*

*1) System dynamics:* Combining eqs. (1) and (2), the mass spring damper system with actuator dynamics can be written in state space form as:

$$\begin{bmatrix}\dot{x}\\ \ddot{x}\\ \dot{u}\end{bmatrix} = \begin{bmatrix}0 & 1 & 0\\ \frac{-k}{m} & \frac{-d}{m} & \frac{1}{m}\\ 0 & 0 & -\frac{1}{\tau}\end{bmatrix}\begin{bmatrix}x\\ \dot{x}\\ u\end{bmatrix} + \begin{bmatrix}0\\ 0\\ \frac{1}{\tau}\end{bmatrix}\tilde{u} \tag{4}$$

We elect the parameters $k = 0.8, d = 0.9, m = 1$ and $\tau = 5$. Under the control law in eq. (3), the mass spring damper system is stable for $K < 4.59$, if $K$ is purely positive. Note that, the system is also unstable for $K < -0.8$. However, $K < 0$ switches the controller behavior from regulation (negative feedback, which drives state to zero) into amplification (positive feedback). As we are concerned with the regulation task, we assume $K > 0$ for a valid control law.

*2) Simulations:* We collect synthetic measurement data of the controlled system via stochastic simulations using GRIND[i] for MATLAB (2024b). We express this as a system of equations involving the controller (eq. (3)) and system (eq. (4)). In this configuration, the controller can be seen as an 'internal' self-regulation mechanism of the dynamical system. This controlled system is subject to stochastic forcing via Brownian motion where noise enters the system additively. Additive noise is chosen here to reflect the presence of non-zero (sensor) noise at the trivial equilibrium states (i.e. $x = \dot{x} = 0$), which the controller is seeking to drive the system towards. This produces persistent perturbations, even at equilibrium, that the controller endeavors to reject. Such noisy measurements are prevalent across many engineered systems[51,52].

A batch of 100 simulations are run for each discrete gain value, $K \in \{0.5, 1.0, 1.5, 3.0\}$, on approach to instability ($K > 4.59$). All other system parameters are kept constant. Additive noise of magnitude $\sigma = 0.25$, affecting only $\dot{x}$ and $\ddot{x}$, is injected into the state equation (eq. (4)) as a Wiener process. Each simulation is initialized with a unique random number generator seed and runs for 50 seconds with a time step of 0.1 seconds. For each simulation, the system states (i.e. position, $x$, and velocity, $\dot{x}$) are recorded at every time step to compose the synthetic measurements.

### *F. Indicators of instability*

*1) Disk margin:* We use the disk margin to determine the stability margins of the controlled mass spring damper system. Given a (linear) system model, H(s), and a controller, K(s), expressed in the Laplace domain, the sensitivity function S(s) of the system may be obtained through eq. (5).

$$S(s) = \frac{1}{1 + H(s)K(s)} \tag{5}$$

Subsequently, the disk margin – which accounts for simultaneous changes to the gain and phase margins of the system – can be obtained through eq. (6) for single-input-single-output systems (as is the case for our example mass spring damper system). Seiler et al.[55] provide a detailed tutorial on how to calculate the disk margin, including systems with more inputs and outputs.

$$DM = \frac{1}{\left\| S - \frac{1}{2} \right\|_{\infty}} \tag{6}$$

*2) Backward reachable set:* Though the backward reachable set is invaluable in theory, computing such sets in practice presents a challenge due to the dependence on an accurate system model and computational intractability (when using standard level-set methods) for high-dimensional systems[57]. Nonetheless, we may leverage the closed-form solution of linear autonomous systems (eq. (7)) to easily compute the backward reachable set. We use the state transition matrix, $\tilde{A}$, of the controlled mass spring damper system (eq. (8)) to determine the evolution of state trajectories starting from $x(t_0)$ that terminate after $(T - t_0)$ seconds at $x(T)$.

$$x(T) = e^{\tilde{A}(T-t_0)}x(t_0) \tag{7}$$

---

[i] Freely available here: https://sparcs-center.org/grind/

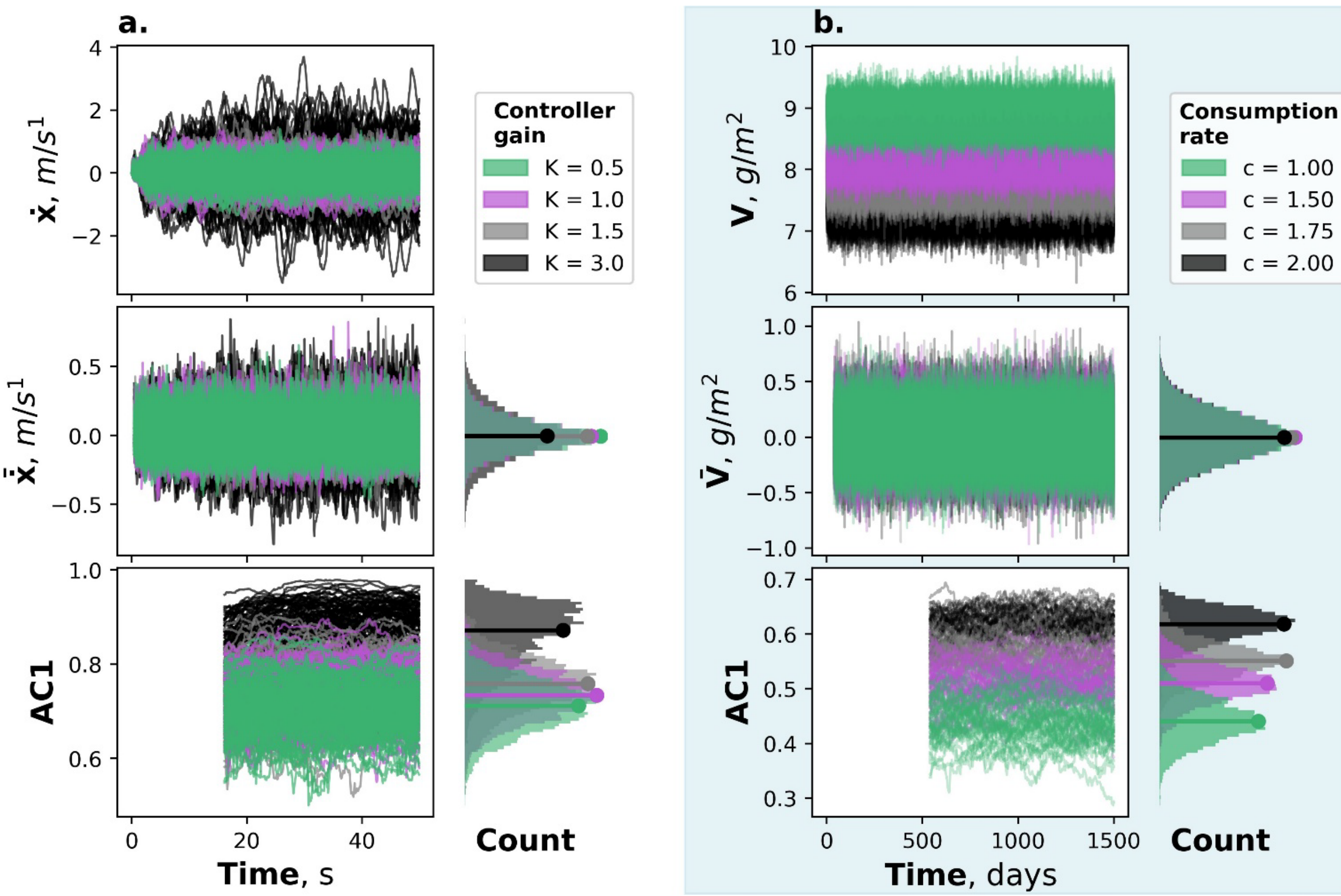

**Figure 9** | **Generic indicators of sudden changes in dynamical system behavior**. Illustrated are the generic indicator construction steps from a state measurement to the lag-1 autocorrelation (AC1) metric of critical slowing down for the controlled mass spring damper system (**a**) and resource exploitation ecological system (**b**). Synthetic measurement data is collected through multiple simulations using GRIND for MATLAB with white noise perturbations. The first row of plots shows the raw measurement signals for different critical parameter values on approach to a sudden shift in dynamical system behavior. Slow trends in these signals are removed via a moving average filter that slides along the measurement data to produce a valid first order autoregressive process. Subsequently, the AC1 is estimated via the Pearson correlation coefficient taken within yet another sliding window over the detrended timeseries. The resultant AC1 histograms depict the characteristic increase in AC1 as the tipping point is approached (from green, to purple, to grey, to black).

$$\tilde{A} = \begin{bmatrix} 0 & 1 & 0 \\ \frac{-k}{m} & \frac{-d}{m} & \frac{1}{m} \\ -K\frac{1}{\tau} & 0 & -\frac{1}{\tau} \end{bmatrix} \tag{8}$$

The backward reachable set is constructed as the set of initial system states (i.e. position, $x$, and velocity, $\dot{x}$) which terminate at the equilibrium point (i.e. $x = \dot{x} = 0$) by the end of the allotted time horizon. For plot **c** in fig. 8, we initialize the system states on a grid and choose $T = 12$ seconds, affording enough time for the system to reach its steady state values. Moreover, the effects of disturbances and noise on the backward reachable set is ignored here since we aim only to give an impression on how the backward reachable set changes as the system progresses towards instability, rather than a robust quantification of its exact bounds.

*3) Critical slowing down:* As with the quadrotor, we rely on the lag-1 autocorrelation (AC1) metric of critical slowing down (CSD) as our generic indicator of instability. Moreover, we compute these indicators for both the position, $x$, and velocity, $\dot{x}$, of the mass spring damper system in order to explicitly show the relation between CSD and the changes in the backward reachable set of the system. Though we incrementally shift the mass spring damper to instability, we nonetheless compute the generic indicators over moving windows sliding along the (synthetic) measurement data. The result is two continuous indicators of instability, one for the position and the other for velocity. Such continuous monitors are appropriate for prompt instability detection in real-world monitoring applications where changes to the system dynamics can occur during operation or between sets of operations (e.g. due to changing environmental conditions).

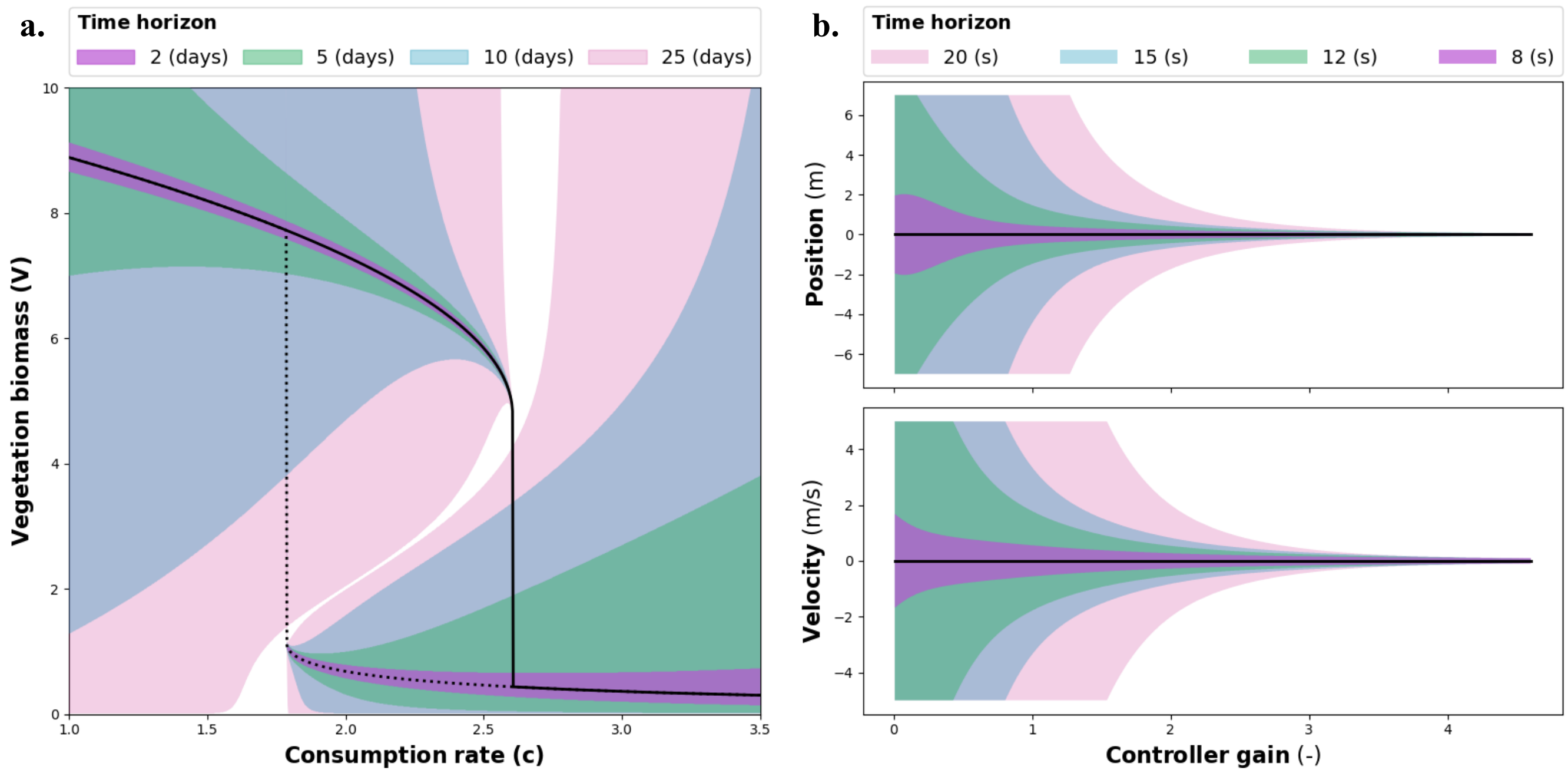

**Figure 10 | Change in backward reachable set as a function of the critical parameter.** The backward reachable set (BRS) is shown for the resource exploitation ecological system in **a**. For reference, the BRS of a controlled mass spring damper system is also shown in **b**. For both systems, a clear contraction in the BRS is observed as the transition points are approached, for all time horizons. The BRS of these systems are computed through deterministic simulations of the system along a grid of critical parameter values. For each critical parameter value, the system states are initialized on a grid, without noise, and are simulated forward in time for the different time horizons. The BRS is composed of the initial states which terminate at the equilibrium states.

To this end, a moving average detrender is used to remove trends from both state variables individually along a moving window sliding across their respective timeseries. An averaging window of 5 samples (equal to 0.5 seconds) results in suitable detrending (i.e. produces an autoregressive process). Subsequently, the AC1 is calculated over a rolling window of 157 samples (equal to 15.7 seconds) running along each detrended timeseries individually. This window size is chosen based on the slowest eigenvalues of the uncontrolled system (i.e. eq. (4) with $\tilde{u} = 0$) such that their dynamics can be observed within the observation window. Within this window, the AC1 coefficient is estimated through a Pearson correlation of the detrended signal with itself lagged by one sample.

Finally, the AC1 values of each simulation at a fixed feedback gain, *K*, are concatenated into a single representative AC1 histogram of all the values observed under *K*. These histograms are depicted in fig. 8 **a**. The entire AC1 construction process, from measurement to AC1 histogram, is illustrated in fig. 9 **a**.

### *G. Control theory for ecological models*

*1) Vegetation over-grazing model:* As an example of an ecological system, we consider the resource exploitation model (eq. (9)) which exhibits critical slowing down (CSD)[54,58]. This model is given by:

$$dV = \left(rV\left(1 - \frac{V}{K}\right) - c\frac{V^2}{V^2 + h^2}\right)dt + \sigma V dW \qquad (9)$$

where the vegetation biomass, $V$, grows logistically with growth rate, *r*, up to a carrying capacity, *K*. The consumption rate of the vegetation is given by *c* with *h* denoting the half-saturation constant. This system is subject to multiplicative white noise, $dW$, with intensity $(\sigma V)^2$ per time step, $dt$. Following[54] we adopt $r = 1, h = 1, K = 10$ and $\sigma = 0.03$.

Though it is well-established that this ecological system experiences CSD[28,54], we nonetheless calculate the generic indicators for this system to highlight the consistency in the behavior of these indicators across both ecological and controlled systems. Here, we construct a synthetic ecological data set through a series of stochastic simulations of eq. (9) for different consumption rates ($c = \{1.00, 1.50, 1.75, 2.00\}$) held fixed throughout a simulation. For each value of *c*, 50 independent simulations are run using GRIND for MATLAB (R2024b), each with a unique pseudo-random number generator seed. Each simulation runs for 1500 days with a time step of $dt = 1$ days.

To mirror the processing steps of the controlled systems studied here, we remove slow trends from the vegetation biomass measurements through a moving average detrender. An averaging window size of 40 days produces a suitable first order autoregressive process. Using this detrended timeseries, the lag-1 autocorrelation (AC1) metric of CSD is calculated over a sliding window of 500 days. Within this 500 day window, the AC1 coefficient is estimated through a Pearson correlation of the detrended vegetation biomass with itself lagged by one sample. This process is illustrated in fig. 9 **b**.

*2) Backward reachability in ecology:* As the bifurcation points of eq. (9) are approached, the basin of attraction of the associated equilibrium points shrinks. This is reflected by a shrinkage in the backward reachable set (BRS) en route to a bifurcation, shown in fig. 10. Also depicted, for reference, is the contracting BRS of the actuated mass spring damper system (eq. (3) and eq. (4)) as it shifts towards instability. For both systems, the BRS in fig. 10 is computed using a system model evaluated on a grid of the system states and critical parameter. For each critical parameter value, the system is initialized, without noise, over a mesh of the states and is simulated[ii] forward for the different time horizons. These deterministic simulations are run in Python (3.13). An initial state qualifies for the BRS if its terminal state is at the equilibrium states by the end of the time horizon (i.e. simulation).

Though the change in morphology of the BRS due to variations in the critical parameter can be non-trivial, such as the 'ghost' (white) regions in fig. 10 **a**, the shrinkage of the set alone can be indicative of an approaching critical transition. To this end, such reachable sets may be used alongside a time horizon to help design (conservation) policies.

When afforded the luxury of representative models of an (ecological) system, the BRS can give temporal insights into policy design. For example, how long before an ecological system can be expected to return to equilibrium and, if this time frame is too long, what 'controllers' (i.e. conservation policies) can be followed to expedite this recovery. Contrarily, if time is not a concern, what strategies are least intrusive, most resilient, or most convenient to implement? This can help motivate the trade-off between different conservation policies. In a similar capacity, the BRS can be used to determine viable, perhaps even counter-intuitive, paths to recovery for multi-state systems through the gradients afforded by the BRS.

## Supplementary References

[ii] Note that there are more effective ways to determine the BRS of a system than via simulations.